\documentclass[10pt,twocolumn,letterpaper]{article}

\usepackage{iccv}
\usepackage{times}
\usepackage{epsfig}
\usepackage{graphicx}
\usepackage{amsmath}
\usepackage{amssymb}

\usepackage{color}
\usepackage{times}
\usepackage{epsfig}
\usepackage{booktabs}
\usepackage{pifont}
\usepackage{url}
\usepackage{subcaption}
\usepackage[T1]{fontenc}
\usepackage{resizegather}
\usepackage{comment}
\usepackage{authblk}

 \normalsize

\usepackage[pagebackref=true,breaklinks=true,letterpaper=true,colorlinks,bookmarks=false]{hyperref}

 \iccvfinalcopy 

\ificcvfinal\fi
\begin{document}

\title{Robust Optimization for Deep Regression}

\author[1,2]{Vasileios Belagiannis}
\author[1,3]{Christian Rupprecht}
\author[4]{Gustavo Carneiro}
\author[1,3]{Nassir Navab}
\affil[1]{Computer Aided Medical Procedures, Technische Universit\"at M\"unchen}
\affil[2]{Visual Geometry Group, Department of Engineering Science, University of Oxford}
\affil[3]{Johns Hopkins University}
\affil[4]{Australian Centre for Visual Technologies, University of Adelaide}
\affil[ ]{\normalsize \textit{vb@robots.ox.ac.uk}, \normalsize \textit {\{christian.rupprecht, navab\}@in.tum.de}, \normalsize \textit{gustavo.carneiro@adelaide.edu.au}}

\maketitle
\thispagestyle{empty}

\begin{abstract}
Convolutional Neural Networks (ConvNets) have successfully contributed to improve the accuracy of regression-based methods for computer vision tasks such as human pose estimation, landmark localization, and object detection. The network optimization has been usually performed with L2 loss and without considering the impact of outliers on the training process, where an outlier in this context is defined by a sample estimation that lies at an abnormal distance from the other training sample estimations in the objective space. In this work, we propose a regression model with ConvNets that achieves robustness to such outliers by minimizing Tukey's biweight function, an M-estimator robust to outliers, as the loss function for the ConvNet. In addition to the robust loss, we introduce a coarse-to-fine model, which processes input images of progressively higher resolutions for improving the accuracy of the regressed values. In our experiments, we demonstrate faster convergence and better generalization of our robust loss function for the tasks of human pose estimation and age estimation from face images. We also show that the combination of the robust loss function with the coarse-to-fine model produces comparable or better results than current state-of-the-art approaches in four publicly available human pose estimation datasets.
\end{abstract}

\section{Introduction}

\setlength{\tabcolsep}{0.3pt}
\begin{figure}[t]
\centering
\begin{tabular}{c}
\includegraphics[scale=0.115, angle=0]{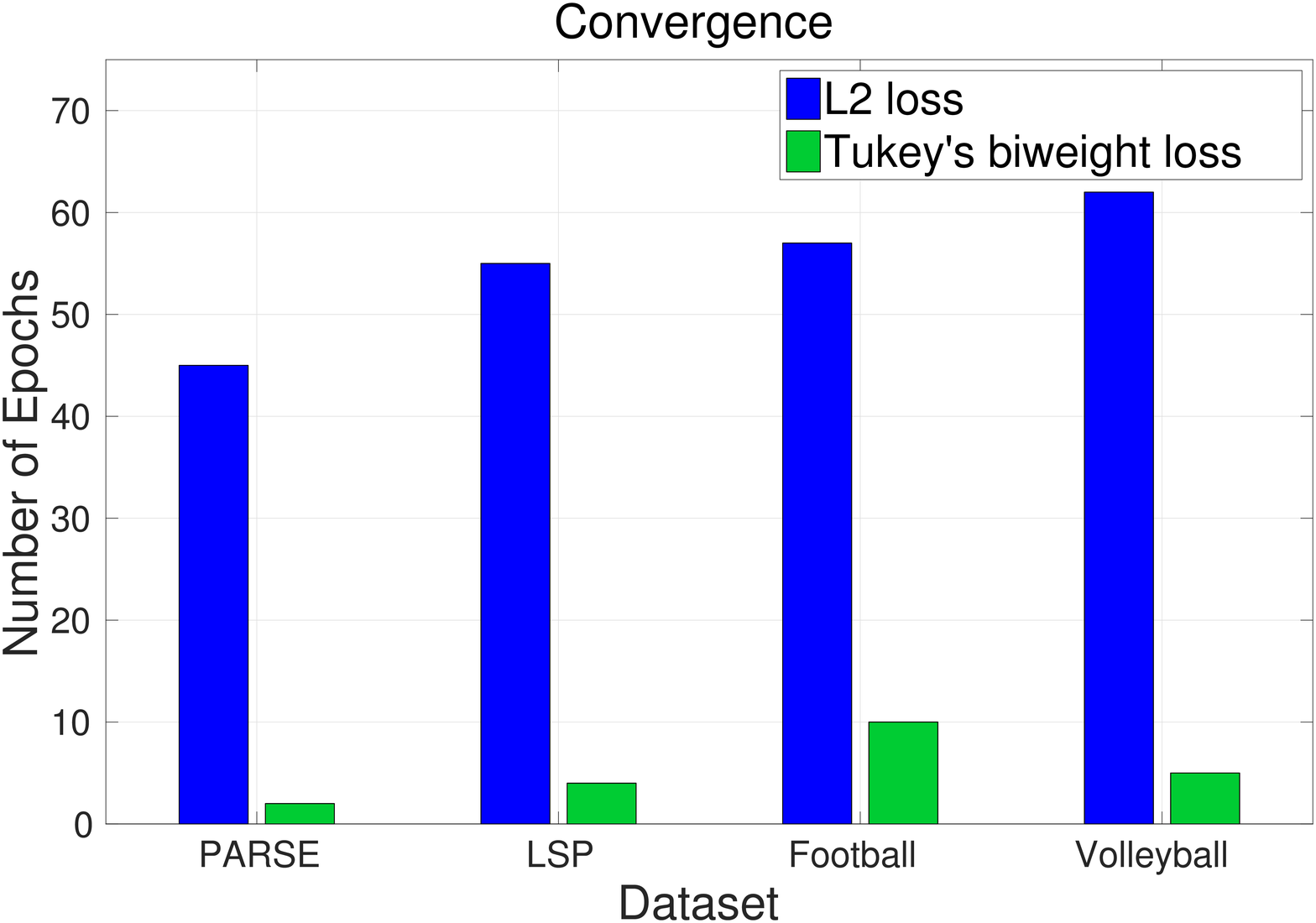} \label{fig:convergence}\\
\includegraphics[scale=0.115, angle=0]{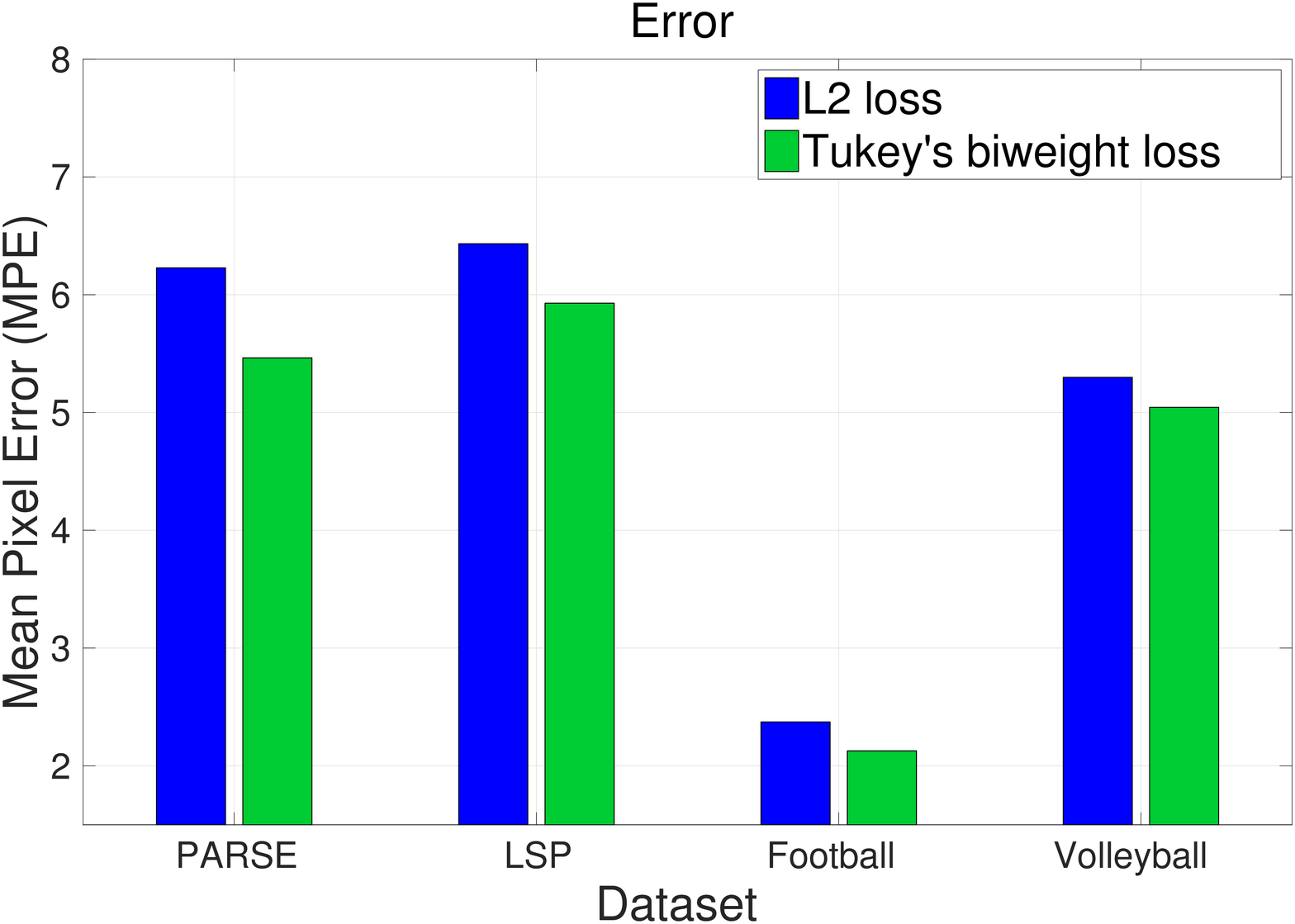} \label{fig:crossLSP}
\end{tabular}
\caption{\small {\bf Comparison of $L2$ and $Tukey's$ $biweight$ loss functions}: We compare our results (Tukey's biweight loss) with the standard $L2$ loss function on the problem of 2D human pose estimation (PARSE~\cite{yang2013articulated}, LSP~\cite{Johnson10}, Football~\cite{kazemi2013multi} and Volleyball~\cite{belagiannis2014holistic} datasets). On top, the convergence of $L2$ and Tukey's biweight loss functions is presented, while on the bottom, the graph shows the mean pixel error (MPE) comparison for the two loss functions. For the convergence computation, we choose as reference error, the smallest error using $L2$ loss (blue bars in bottom graph). Then, we look for the epoch with the closest error in the training using Tukey's biweight loss function.}
\label{fig:resultsComparison}
\end{figure}

Deep learning has played an important role in the computer vision field in the last few years. In particular, several methods have been proposed for challenging tasks, such as classification~\cite{krizhevsky2012imagenet}, detection~\cite{girshick2014rich}, categorization \cite{zhang2014part}, segmentation \cite{long2015fully}, feature extraction \cite{sermanet2013overfeat} and pose estimation~\cite{chen2014articulated}. State-of-the-art results in these tasks have been achieved with the use of Convolutional Neural Networks (ConvNets) trained with  backpropagation~\cite{lecun1989backpropagation}. Moreover, the majority of the tasks above are defined as classification problems, where the ConvNet is trained to minimize a softmax loss function~\cite{chen2014articulated, krizhevsky2012imagenet}. Besides classification, ConvNets have been also trained for regression tasks such as human pose estimation \cite{li_3Ddeep, toshev2014deeppose}, object detection~\cite{szegedy2013deep}, facial landmark detection~\cite{sun2013deep} and depth prediction~\cite{eigen2014depth}. In regression problems, the training procedure usually optimizes an $L2$ loss function plus a regularization term, where the goal is to minimize the squared difference between the estimated values of the network and the ground-truth. However, it is generally known that $L2$ norm minimization is sensitive to outliers, which can result in poor generalization depending on the amount of outliers present during training~\cite{huber2011robust}. Without loss of generality, we assume that the samples are drawn from an unknown distribution and outliers are sample estimations that lie at an abnormal distance from other training samples in the objective space~\cite{moore1989introduction}. Within our context, outliers are typically represented by uncommon samples that are rarely encountered in the training data, such as  rare body poses in human pose estimation, unlikely facial point positions in facial landmark detection or samples with imprecise ground-truth annotation. In the presence of outliers, the main issue of using $L2$ loss in regression problems is that outliers can have a disproportionally high weight and consequently influence the training procedure by reducing the generalization ability and increasing the convergence time.

In this work, we propose a loss function that is robust to outliers for training ConvNet regressors. Our motivation originates from Robust Statistics, where the problem of outliers has been extensively studied over the past decades, and several robust estimators have been proposed for reducing the influence of outliers in the model fitting process~\cite{huber2011robust}.  Particularly in a ConvNet model, a robust estimator can be used in the loss function minimization, where training samples with unusually large errors are downweighted such that they minimally influence the training procedure. It is worth noting that the training sample weighting provided by the robust estimator is done without any hard threshold between inliers and outliers. Furthermore, weighting training samples also conforms with the idea of curriculum~\cite{bengio2009curriculum} and self-paced learning~\cite{kumar2010self}, where each training sample has different contribution to the minimization depending on its error. Nevertheless, the advantage in the use of a robust estimator, over the concept of curriculum or self-paced learning, is that the minimization and weighting are integrated in a single function. 

We argue that training a ConvNet using a loss function that is robust to outliers results in faster convergence and better generalization (Fig.~\ref{fig:resultsComparison}). We propose the use of \textit{Tukey's biweight} function, a robust M-estimator, as the loss function for the ConvNet training in regression problems (Fig.~\ref{fig:tukeyFig}). Tukey's biweight loss function weights the training samples based on their residuals (notice that we use the terms residual and error interchangeably, even if the two terms are not identical, with both standing for the difference between the true and estimated values). Specifically, samples with unusually large residuals (i.e.~outliers) are downweighted and consequently have small influence on the training procedure. Similarly, inliers with insignificant residuals are also downweighted in order to prevent instabilities around local minima. Therefore, samples with residuals that are not too high or too small (i.e.~inliers with significant residuals) have the largest influence on the training procedure. In our ConvNet training, this influence is represented by the gradient magnitude of Tukey's biweight loss function, where in the backward step of backpropagation, the gradient magnitude of the outliers is low, while the gradient magnitude of the inliers is high except for the ones close to the local minimum. In Tukey's biweight loss function, there is no need to define a hard threshold between inliers and outliers. It only requires a tuning constant for suppressing the residuals of the outliers. We normalize the residuals with the median absolute deviation ($\operatorname{MAD}$)~\cite{venables2002modern}, a robust approximation of variability, in order to preassign the tuning constant and consequently be free of parameters.

To demonstrate the advances of Tukey's biweight loss function, we apply our method to 2D human pose estimation in still images and age estimation from face images. In human pose estimation, we propose a novel coarse-to-fine model to improve the accuracy of the localized body skeleton, where the first stage of the model is based on an estimation of all output variables using the input image, and the second stage relies on an estimation of different subsets of the output variables using higher resolution input image regions extracted using the results of the first stage. In the experiments, we evaluate our method on four publicly available human pose datasets (PARSE~\cite{yang2013articulated}, LSP~\cite{Johnson10}, Football~\cite{kazemi2013multi} and Volleyball~\cite{belagiannis2014holistic}) and one on age estimation~\cite{escalera2015chalearn} in order to show that: 1.~the proposed robust loss function allows for faster convergence and better generalization compared to the $L2$ loss; and 2.~the proposed coarse-to-fine model produces comparable to better results than the state-of-the-art for the task of human pose estimation.

\section{Related Work}
In this section, we discuss deep learning approaches for regression-based computer vision problems. In addition, we review the related work on human pose estimation, since it comprises the main evaluation of our method. We refer to~\cite{schmidhuber2015deep} for an extended overview of deep learning and its evolution.

\paragraph{Regression-based deep learning.}  A large number of regression-based deep learning algorithms have been recently proposed, where the goal is to predict a set of interdependent continuous values. For instance, in object and text detection, the regressed values correspond to a bounding box for localisation~\cite{jaderberg15reading, szegedy2013deep}, in human pose estimation, the values represent the positions of the body joints on the image plane~\cite{li_3Ddeep, Pfister14a, toshev2014deeppose}, and in facial landmark detection, the predicted values denote the image locations of the facial points~\cite{sun2013deep}. In all these problems, a ConvNet has been trained using an $L2$ loss function, without considering its vulnerability to outliers. It is interesting to note that some deep learning based regression methods combine the $L2$-based objective function with a classification function, which effectively results in a regularization of $L2$ and increases its robustness to outliers. For example, Zhang et al.~\cite{zhang2014facial} introduce a ConvNet that is optimized for landmark detection and attribute classification, and they show that the combination of softmax and $L2$ loss functions improves the network performance when compared to the minimization of $L2$ loss alone.
Wang et al.~\cite{wang2014deep} use a similar strategy for the task of object detection, where they combine the bounding box localization (using an $L2$ norm) with object segmentation. The regularization of the $L2$ loss function has been also addressed by Gkioxari et al.~\cite{gkioxari2014r}, where the function being minimized comprises a body pose estimation term (based on $L2$ norm) and an action detection term.  Finally, other methods have also been proposed to improve the robustness of the $L2$ loss to outliers, such as the use of complex objective functions in depth estimation~\cite{eigen2014depth} or multiple $L2$ loss functions for object generation~\cite{DB14a}. However, to the best of our knowledge, none of the proposed deep learning approaches handles directly the presence of outliers during training with the use of a robust loss function, like we propose in this paper. Robust estimation methods, within our context, can be found in the literature for training artificial neural networks~\cite{neuneier1998train} or Hopfield-Tank networks~\cite{darrell1990segmentation}, but not for deep networks. For instance, a smoother function than $L2$, using a \textit{logcosh} loss, has been proposed in~\cite{neuneier1998train} or a Conditional Density Estimation Network (CDEN) in~\cite{neuneier1994estimation}.

\begin{figure}
\centering
\begin{tabular}{ccccc}
\includegraphics[scale=0.125, angle=-0]{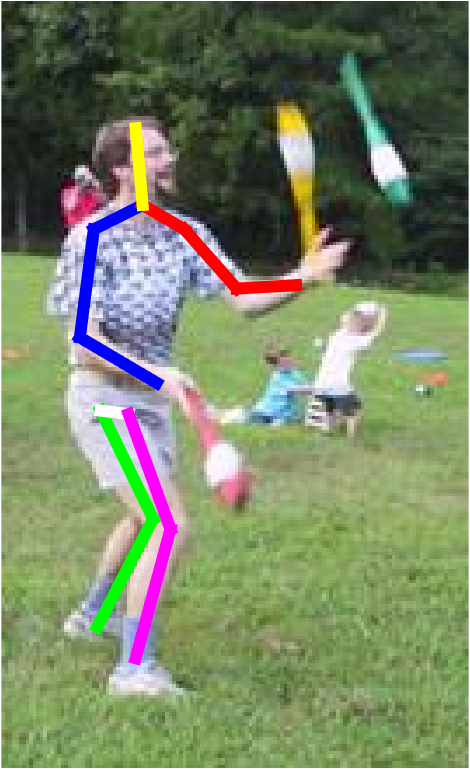}&
\includegraphics[scale=0.125, angle=-0]{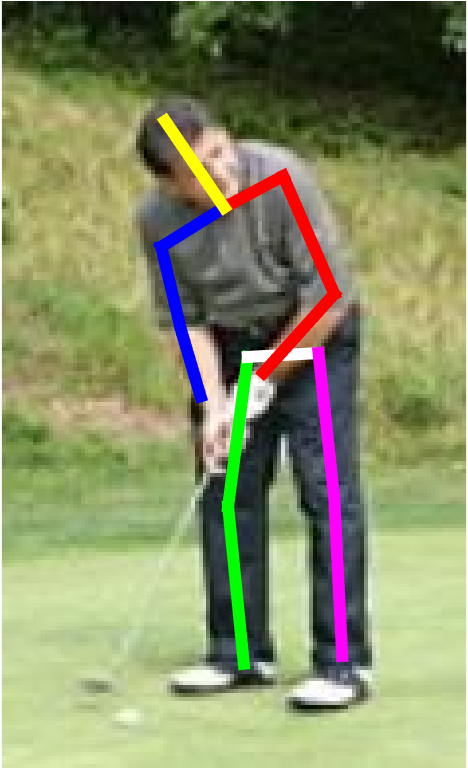}&
\includegraphics[scale=0.125, angle=-0]{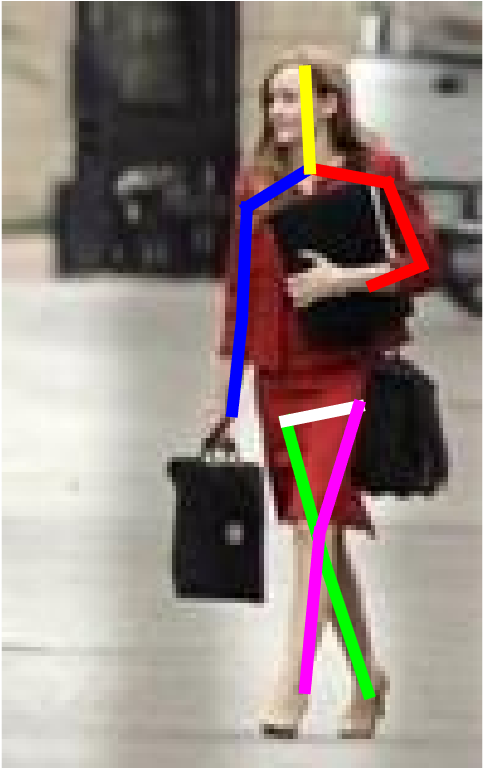}&
\includegraphics[scale=0.13, angle=-0]{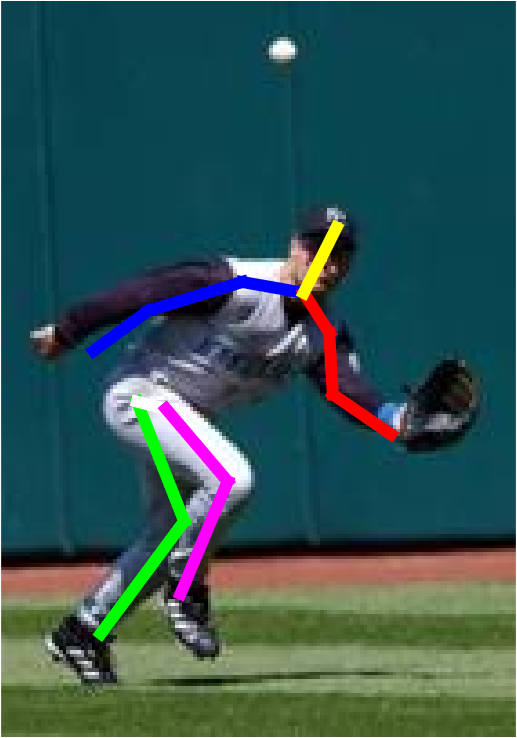}&
\includegraphics[scale=0.155, angle=0]{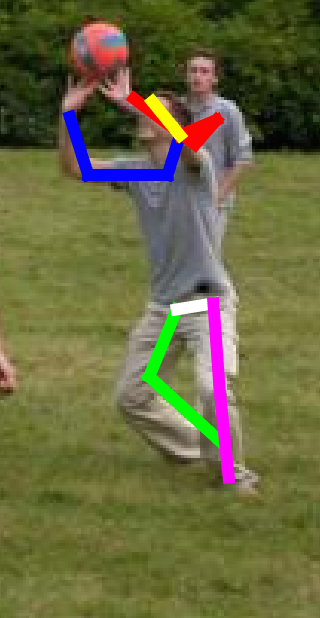}\\
\end{tabular}
\caption{\small {\bf Our Results} Our results on 2D human pose estimation on the PARSE \cite{yang2013articulated} dataset.}
\label{fig:resultsImgParse}
\end{figure}

\paragraph{Human pose estimation} The problem of human pose estimation from images can be addressed by regressing a set of body joint positions. It has been extensively studied from the single- and multi-view perspective, where the standard ways to tackle the problem are based on part-based models~\cite{andriluka2009pictorial, belagiannis20143d, Ferrari08, pishchulin2012articulated, sigal2012loose, yang2013articulated} and holistic approaches~\cite{IonescuSminchisescu11, gavrila2007bayesian, mori2002estimating}. Most of the recent proposals using deep learning approaches have extended both part-based and holistic models. In part-based models, the body is decomposed into a set of parts and the goal is to infer the correct body configuration from the observation. The problem is usually formulated using a conditional random field (CRF), where the unary potential functions include, for example, body part classifiers, and the pairwise potential functions are based on a body prior. Recently, part-based models have been combined with deep learning for 2D human pose estimation~\cite{chen2014articulated, ouyang2014multi, tompson2014joint}, where deep part detectors serve as unary potential functions and also as image-based body prior for the computation of the pairwise potential functions. Unlike part-based models, holistic pose estimation approaches directly map image features to body poses~\cite{gavrila2007bayesian, mori2002estimating}. Nevertheless, this mapping has been shown to be a complex task, which ultimately produced less competitive results when compared to part-based models. Holistic approaches have been re-visited due to the recent advances in the automatic extraction of high level features using ConvNets~\cite{li_3Ddeep, Pfister14a, toshev2014deeppose}. More specifically, Toshev et al.~\cite{toshev2014deeppose} have proposed  a cascade of ConvNets for 2D human pose estimation in still images. Furthermore, temporal information has been included to the ConvNet training for more accurate 2D body pose estimation~\cite{Pfister14a} and the use of ConvNets for 3D body pose estimation from a single image has also been demonstrated in \cite{li_3Ddeep}.  Nevertheless, these deep learning methods do not address the issue of the presence of outliers in the training set.

The main contribution of our work is the introduction of Tukey's biweight loss function for regression problems based on ConvNets. We focus on 2D human pose estimation from still images (Fig.~\ref{fig:resultsImgParse}), and as a result our method can be classified as a holistic approach and is close to the cascade of ConvNets from~\cite{toshev2014deeppose}. However, we optimize a robust loss function instead of the $L2$ loss of~\cite{toshev2014deeppose} and empirically show that this loss function leads to more efficient training (i.e faster convergence) and better generalization results. 
\section{Robust Deep Regression}

In this section, we introduce the proposed robust loss function for training ConvNets on regression problems. Inspired by M-estimators from Robust Statistics~\cite{black1996unification}, we propose the use of Tukey's biweight function as the loss to be to be minimized during the network training.

The input to the network is an image $\mathbf{x}:\Omega \rightarrow \mathbb R$ and the output is a real-valued vector $\mathbf{y}=(y_{1}, y_{2},\dots, y_{N})$ of $N$ elements, with $y_i \in \mathbb{R}$. Given a training dataset $\{ (\mathbf{x}_{s}, \mathbf{y}_{s})\}_{s=1}^{S}$ of $S$ samples, our goal is the training of a ConvNet, represented by the function $\phi(.)$, under the minimization of Tukey's biweight loss function with  backpropagation~\cite{rumelhart1988learning} and stochastic gradient descent~\cite{bottou2010large}. This training process produces a ConvNet with learnt parameters $\boldsymbol{\theta}$ that is effectively a mapping between the input image $\mathbf{x}$ and output $\mathbf{y}$, represented by:
\begin{equation}\label{ConvNetEq}
\hat{\mathbf{y}}=\phi ( \mathbf{x}; \boldsymbol{\theta}),
\end{equation}

where $\hat{\mathbf{y}}$ is the estimated output vector. Next, we present the architecture of the network, followed by Tukey's biweight loss function. In addition, we introduce a coarse-to-fine model for capturing features in different image resolutions for improving the accuracy of the regressed values.

\setlength{\tabcolsep}{15pt}
\begin{figure*}[t]
\centering
\begin{tabular}{cc}
\includegraphics[scale=0.41]{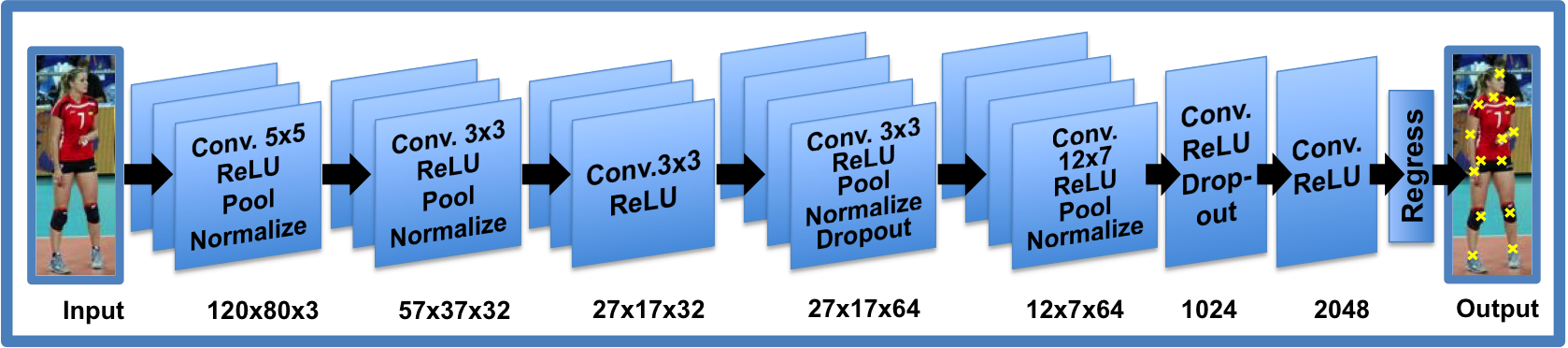} \label{fig:net1}&
\includegraphics[scale=0.41]{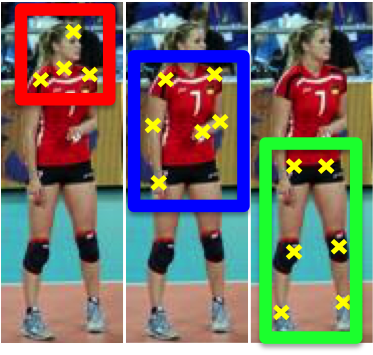} \label{fig:cascade}\\
{\small Network Architecture} & {\small Coarse-to-Fine Model} \\
\end{tabular}
\caption{\small {\bf Network and cascade structure}: Our network consists of five convolutional layers, followed by two fully connected layers. We use relative small kernels for the first two layers of convolution due to the smaller input image in comparison to~\cite{krizhevsky2012imagenet}. Moreover, we use a small number of filters because we have observed that regression tasks required fewer features than classification~ \cite{krizhevsky2012imagenet}.  The last three images (Coarse-to-Fine Model) on the right show the $C=3$ image regions and respective subsets of $\hat{\mathbf{y}}$ used by the cascade of ConvNets in the proposed coarse-to-fine model. }
\label{fig:network}
\end{figure*}

\subsection{Convolutional Neural Network Architecture}
\label{sec:CNN}

Our network takes as input an RGB image and regresses a $N$-dimensional vector of continuous values. As it is presented  in Fig.~\ref{fig:network}, the architecture of the network consists of five convolutional layers, followed by two fully connected layers and the output that represents the regressed values. The structure of our network is similar to Krizhevsky's~\cite{krizhevsky2012imagenet}, but we use smaller kernels and fewer filters in the convolutional layers. Our fully connected layers are smaller as well, but as we demonstrate in the experimental section, the smaller number of parameters is sufficient for the regression tasks considered in this paper. In addition, we apply local contrast normalization, as proposed in~\cite{krizhevsky2012imagenet}, before every convolutional layers and max-pooling after each convolutional layer in order to reduce the image size. We argue that the benefits of max-pooling, in terms of reducing the computational cost, outweighs the potential negative effect in the output accuracy for regression problems. Moreover, we use dropout~\cite{srivastava2014dropout} in the fourth convolutional and first fully connected layers to prevent overfitting. The activation function for each layer is the rectified linear unit (ReLU)~\cite{nair2010rectified}, except for the last layer, which uses a linear activation function for the regression. Finally, we use our robust loss function for training the network of Fig.~\ref{fig:network}.

\subsection{Robust Loss Function}
\label{sec:robustLoss}
The training process of the ConvNet is accomplished through the minimization of a loss function that measures the error between ground-truth and estimated values (i.e.~the residual).  In regression problems, the typical loss function used is the $L2$ norm of the residual, which during backpropagation produces a gradient whose magnitude is linearly proportional to this difference.  This means that estimated values that are close to the ground-truth (i.e.~inliers) have little influence during backpropagation, but on the other hand, estimated values that are far from the ground-truth (i.e.~outliers) can bias the whole training process given the high magnitude of their gradient, and as a result adapt the \mbox{ConvNet} to these outliers while deteriorating its performance for the inliers. Recall that we consider the outliers to be estimations from training samples that lie at an abnormal distance from other sample estimations in the objective space. This is a classic problem addressed by Robust Statistics~\cite{black1996unification}, which is solved with the use of a loss function that weights the training samples based on the residual magnitude. The main idea is to have a loss function that has low values for small residuals, and then usually grows linearly or quadratically for larger residuals up to a point when it saturates. This means that only relatively small residuals (i.e.~inliers) can influence the training process, making it robust to the outliers that are mentioned above.

\begin{figure}[h]
\centering
\def\arraystretch{1.5}
\begin{tabular}{cc}
\multicolumn{2}{c}{{\small \textit{Tukey's} biweight loss function and the derivative}}\\
\includegraphics[scale=0.06]{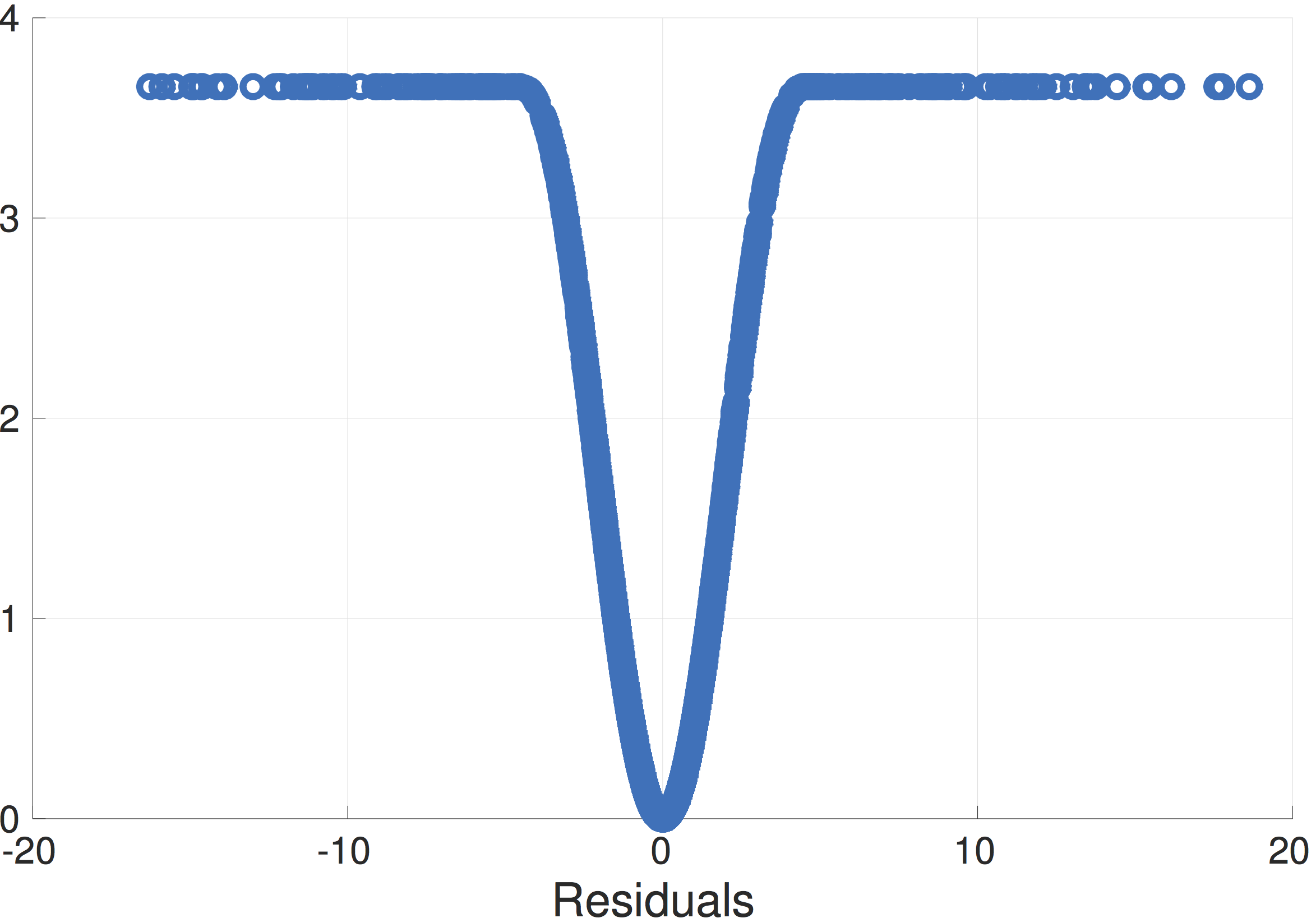} \label{fig:tu1}&
\includegraphics[scale=0.06]{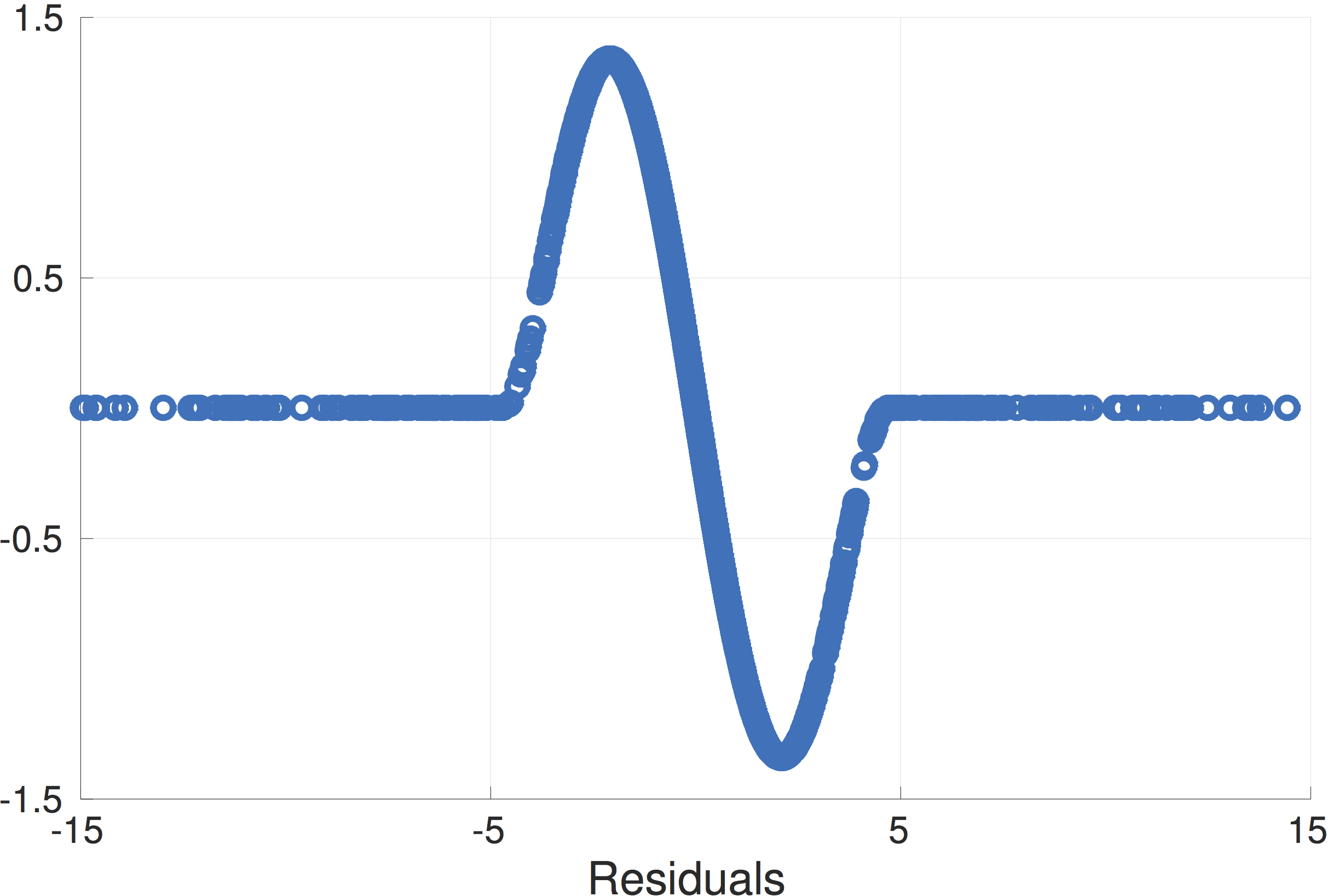} \label{fig:tu2}\\
\end{tabular}
\caption{\small {\bf Tukey's biweight loss function}: Tukey's biweight loss function (left) and its derivative (right) as a function of the training sample residuals.}
\label{fig:tukeyFig}
\end{figure}

There are many robust loss functions that could be used, but we focus on \textit{Tukey's} biweight function~\cite{black1996unification} because of its property of suppressing the influence of outliers during backpropagation (Fig.~\ref{fig:tukeyFig}) by reducing the magnitude of their gradient close to zero.  Another interesting property of this loss function is the soft constraints that it imposes between inliers and outliers without the need of setting a hard threshold on the residuals.  Formally, we define a residual of the $i^{th}$ value of vector $\mathbf{y}$ by:
\begin{equation}\label{Residu}
r_{i}=y_{i}-\hat{y}_{i},
\end{equation}
where $\hat{y}_{i}$ represents the estimated value for the $i^{th}$ value of $\mathbf{y}$, produced by the ConvNet. Given the residual $r_{i}$, Tukey's biweight loss function is defined as:
\begin{equation}\label{Loss}
\rho(r_{i}) = \begin{cases} \frac{c^{2}}{6}\left [ 1-(1-(\frac{r_{i}}{c})^{2} )^{3}   \right ] &\mbox{, if } \left | r_{i} \right |\leq c \\ 
\frac{c^{2}}{6} & \mbox{, otherwise} \end{cases},
\end{equation}
where $c$ is a tuning constant, which if is set to $c=4.6851$, gives approximately $95\%$ asymptotic efficiency as $L2$ minimization on the standard normal distribution of residuals. However, this claim stands for residuals drawn from a distribution with unit variance, which is an assumption that does not hold in general. Thus, we approximate a robust measure of variability from our training data in order to scale the residuals by computing the median absolute deviation ($\operatorname{MAD}$)~\cite{huber2011robust}. $\operatorname{MAD}$ measures the variability in the training data and is estimated as:
\begin{equation}\label{Mad}
\operatorname{MAD}_i = \operatorname*{median}_{k \in \{1,\ldots,S\}}\left(\ \left| r_{i,k} - \operatorname*{median}_{j \in \{1,\ldots,S\}} (r_{i,j}) \right|\ \right),
\end{equation}
for $i \in \{1,...,N\}$ and the subscripts $k$ and $j$ index the training samples. The $\operatorname{MAD}_i$ estimate acts as a scale parameter on the residuals for obtaining unit variance. By integrating $\operatorname{MAD}_i$ to the residuals, we obtain: 
\begin{equation}\label{rMAD}
r_{i}^{\operatorname{MAD}}=\frac{y_{i}-\hat{y}_{i}}{1.4826 \times \operatorname{MAD}_{i}},
\end{equation}
where we scale $\operatorname{MAD}_{i}$ by $1.4826$ in order to make $\operatorname{MAD}_i$ an asymptotically consistent estimator for the estimation of the standard deviation~\cite{huber2011robust}. Then, the scaled residual $r_{i}^{\operatorname{MAD}}$ in Eq.~\eqref{rMAD} can be directly used by Tukey's  biweight loss function Eq.~\eqref{Loss}. We fix the tuning constant based on $\operatorname{MAD}$ scaling and thus our loss function is free of parameters. The final objective function based on Tukey's loss function and $\operatorname{MAD}_i$ estimate is given by:
\begin{equation}\label{Objective}
E = \frac{1}{S}\sum_{s=1}^{S}\sum_{i=1}^{N}\rho\left ( r_{i,s}^{\operatorname{MAD}} \right ).
\end{equation}

We illustrate the functionality of Tukey's biweight loss function in Fig.~\ref{fig:tukeyFig}, which shows the loss function and its derivative as a function of sample residuals in a specific training problem. This is an instance of the training for the LSP \cite{Johnson10} dataset that is further explained in the experiments.

\subsection{Coarse-to-Fine Model}
\label{sec:modelRef}

We adopt a coarse-to-fine model, where initially a single network $\phi(.)$ of Eq.~\eqref{ConvNetEq} is trained from the input images to regress all $N$ values of $\hat{\mathbf{y}}$, and then separate networks are trained to regress subsets of $\hat{\mathbf{y}}$ using the output of the single network $\phi(.)$ and higher resolution input images. Effectively, the coarse-to-fine model produces a cascade of ConvNets, where the goal is to capture different sets of features in high resolution input images, and consequently improve the accuracy of the regressed values. Similar approaches have been adopted by other works~\cite{eigen2014depth, tompson2014joint, toshev2014deeppose} and shown to improve the accuracy of the regression. Most of these approaches refine each element of $\hat{\mathbf{y}}$ independently, while we employ a different strategy of refining subsets of $\hat{\mathbf{y}}$. We argue that our approach constrains the search space more and thus facilitates the optimization.

More specifically, we define $C$ image regions and subsets of $\hat{\mathbf{y}}$ that are included in theses regions (Fig. \ref{fig:network}). Each image region $\mathbf{x}^{c}$, where $c \in \{1,...,C\}$, is cropped from the original image $\mathbf{x}$ based on the output of the single \mbox{ConvNet} of Eq.~\eqref{ConvNetEq}. Then the respective subset of $\hat{\mathbf{y}}$ that falls in the image region $c$ is transformed to the coordinate system of this region. To define a meaningful set of regions, we rely on the specific regression task. For instance, in 2D human pose estimation, the regions can be defined based on the body anatomy (e.g. head and torso or left arm and shoulder);  similarly, in facial landmark localization the regions can be defined based on the face structure (e.g. nose and mouth). This results in training $C$ additional ConvNets $\{ \phi^{c}(.) \}_{c=1}^C$ whose input is defined by the output of the single ConvNet $\phi(.)$ of Eq.~\eqref{ConvNetEq}. The refined output values from the cascade of ConvNets are obtained by:
\begin{equation}\label{refinedY}
\hat{\mathbf{y}}_{ref}=\operatorname{diag}(\mathbf{z})^{-1} \sum_{c=1}^{C}\phi^{c} \left( \mathbf{x}^{c}; \boldsymbol{\theta}^{c}, \hat{\mathbf{y}}(l^{c}) \right ),
\end{equation}
where $l^{c} \subset \{  1,2, \ldots, N \}$ indexes the subset $c$ of $\hat{\mathbf{y}}$, the vector $\mathbf{z} \in \mathbb{N}^N$ has the number of subsets in which each element of $\hat{\mathbf{y}}$ is included and $\boldsymbol{\theta}^{c}$ are the learnt parameters. Every ConvNet of the cascade regresses values only for the dedicated subset $l^{c}$, while its output is zero for the other elements of $\hat{\mathbf{y}}$. To train the ConvNets $\{ \phi^{c}(.) \}_{c=1}^C$ of the cascade, we extract the training data based on the output of the single ConvNet $\phi(.)$ of Eq.~\eqref{ConvNetEq}. Moreover, we use the same network structure that is described in Sec.~\ref{sec:CNN} and the same robust loss function of Eq.~\eqref{Objective}. Finally, during inference, the first stage of the cascade uses the single ConvNet $\phi(.)$ to produce $\hat{\mathbf{y}}$, which is refined by the second stage of the cascade with the  ConvNets $\{ \phi^{c}(.) \}_{c=1}^C$ of Eq.~\eqref{refinedY}. The predicted values $\hat{\mathbf{y}}_{ref}$ of the refined regression function are normalized back to the coordinate system of the image $\mathbf{x}$.

\setlength{\tabcolsep}{1.0pt}

\subsection{Training Details}

The input RGB image to the network has resolution $120\times 80$, as it is illustrated in Fig.~\ref{fig:network}. Moreover, the input images are normalized by subtracting the mean image estimated from the training images\footnote{We have also tried the normalization based on the division by the standard deviation of the training data, but we did not notice any particular positive or negative effect in the results.}. We also use data augmentation in order to regularize the training procedure. To that end, each training sample is rotated and flipped (50 times) as well as a small amount of Gaussian noise is added to the ground-truth values $\mathbf{y}$ of the augmented data. Furthermore, the same training data is shared between the first cascade stage for training the single ConvNet $\phi(.)$ and second cascade stage for training the ConvNets $\{ \phi^{c}(.) \}_{c=1}^C$. Finally, the elements of the output vector of each training sample are scaled to the range $\left [  0,1 \right ]$. Concerning the network parameters, the learning rate is set to $0.01$, momentum to $0.9$, dropout to $0.5$ and the batch size to $230$ samples.

The initialisation of the ConvNets' parameters is performed randomly, based on an unbiased Gaussian distriubtion with standard deviation $0.01$, with the result that many outliers can occur at the beginning of training. To prevent this effect that could slow down the training or exclude samples at all from contributing to the network's parameter update, we increase the $\operatorname{MAD}$ values by a factor of $7$ for the first $50$ training iterations (around a quarter of an epoch). Increasing the variability for a few iterations helps the network to quickly reach a more stable state. Note that we have empirically observed that the number of iterations needed for this $\operatorname{MAD}$ adjustment does not play an important role in the whole training process and thus these values are not hard constraints for convergence.

\section{Experiments}

We evaluate Tukey's biweight loss function for the problem of 2D human pose estimation from still images. For that purpose, we have selected four publicly available datasets, namely PARSE~\cite{yang2013articulated}, LSP~\cite{Johnson10}, Football~\cite{kazemi2013multi} and Volleyball~\cite{belagiannis2014holistic}. All four datasets include sufficient amount of data for training the ConvNets, except for PARSE which has only $100$ training images. For that reason, we have merged LSP and PARSE training data, similar to~\cite{Johnson10}, for the evaluation on the PARSE dataset. For the other three datasets, we have used their training data independently. In all cases, we train our model to regress the 2D body skeleton as a set of joints that correspond to pixel coordinates (Fig.~\ref{fig:resultsAdditional}). We assume that each individual is localized within a bounding box with normalized body pose coordinates. Our first assumption holds for all four datasets, since they include cropped images of the individuals, while for the second we have to scale the body pose coordinates in the range $\left [  0,1 \right ]$. Moreover, we introduce one level of cascade using three parallel networks ($C=3$) based on the body anatomy for covering the following body parts: 1) head - shoulders, 2) torso - hands, and 3)  legs (see Fig.~\ref{fig:network}). In the first part of the experiments, a baseline evaluation is presented, where Tukey's biweight and the standard $L2$ loss functions are compared in terms of convergence and generalization. We also present a baseline evaluation on age estimation from face images~\cite{escalera2015chalearn}, in order to the show generalization of our robust loss in different regression tasks. Finally, we compare the results of our proposed coarse-to-fine model with state-of-the-art methodologies in human pose estimation.

\setlength{\tabcolsep}{0.3pt}
\begin{figure}[t]
\centering
\begin{tabular}{cc}
\includegraphics[scale=0.092, angle=-0]{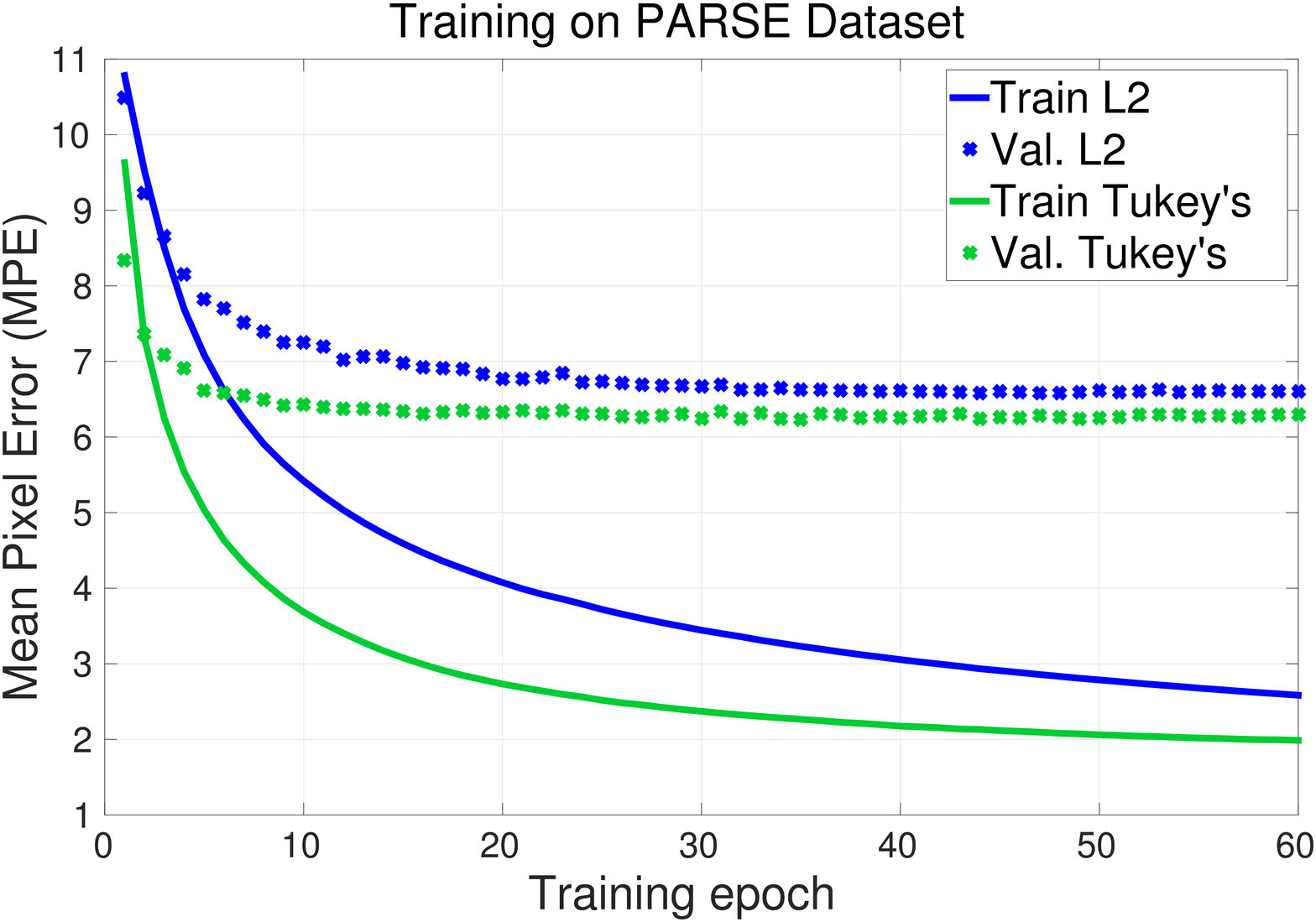} \label{fig:convergence}&
\includegraphics[scale=0.092, angle=-0]{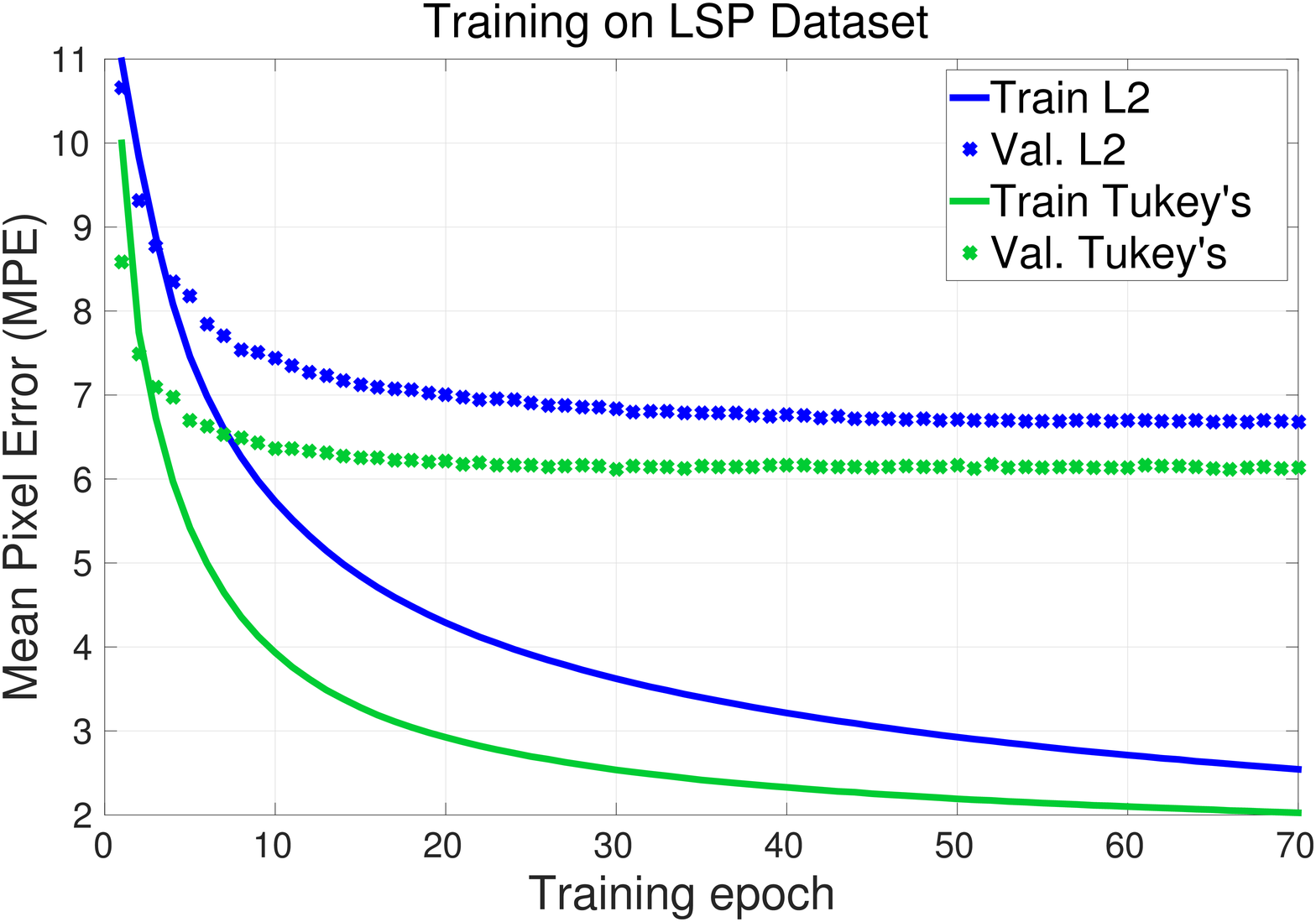} \label{fig:crossLSP}\\
\includegraphics[scale=0.092, angle=-0]{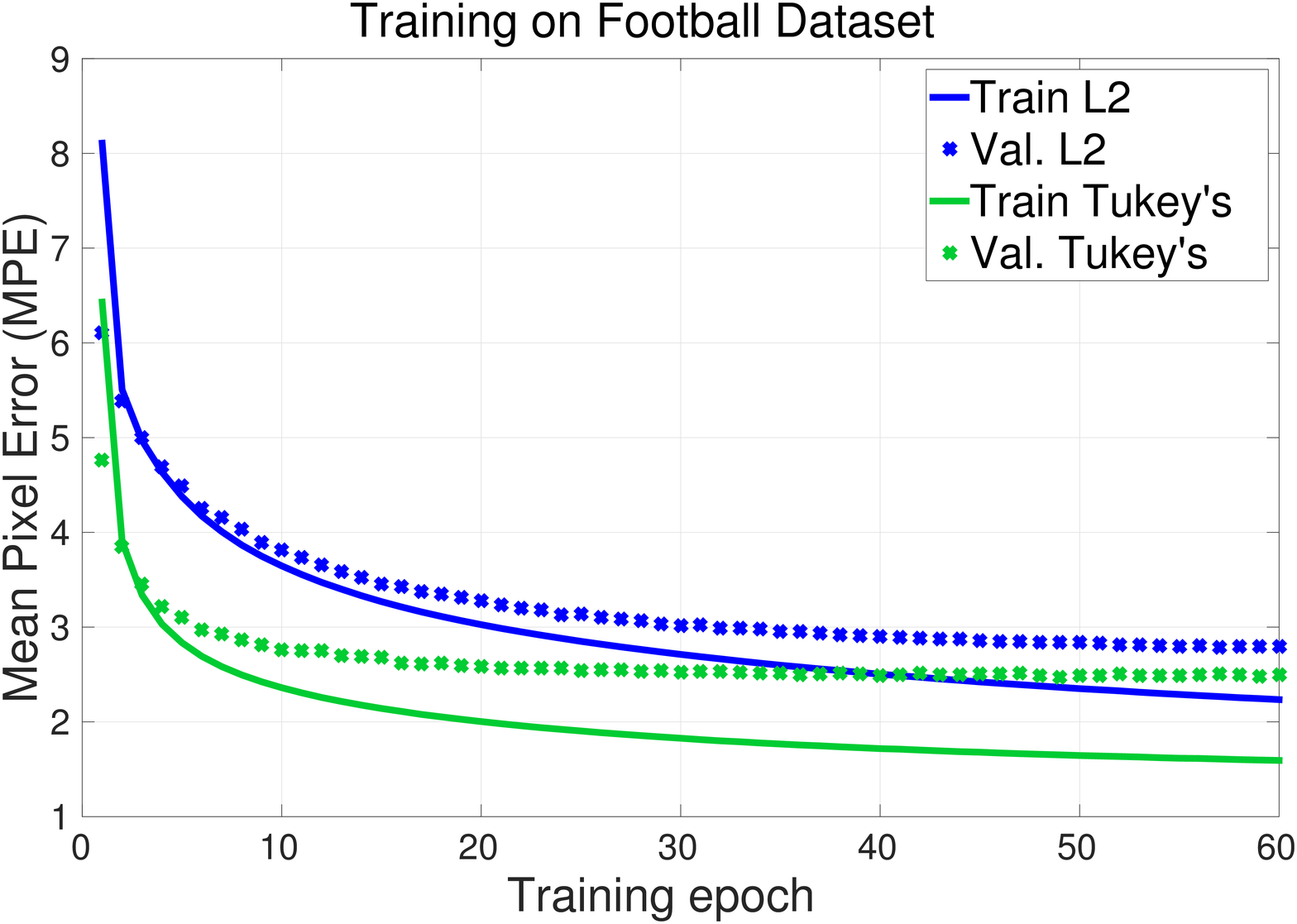} \label{fig:crossFoot}&
\includegraphics[scale=0.092, angle=-0]{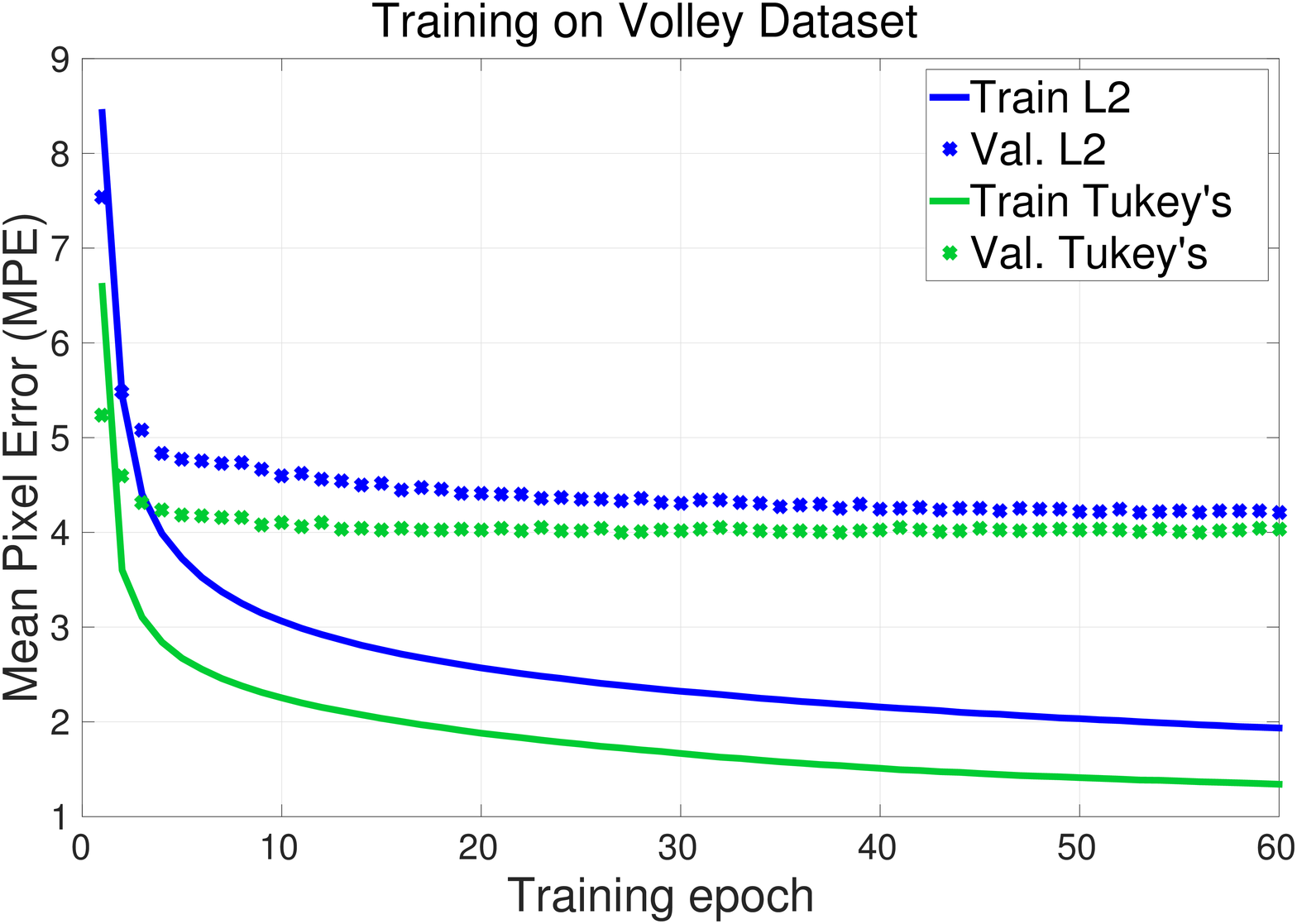} \label{fig:crossVolley}
\\
\end{tabular}
\caption{\small {\bf Comparison of $L2$ and $Tukey's$ $biweight$ loss functions}:In all datasets (PARSE~\cite{yang2013articulated}, LSP~\cite{Johnson10}, Football~\cite{kazemi2013multi} and Volleyball~\cite{belagiannis2014holistic}), $Tukey's$ $biweight$ loss function shows, on average, faster convergence and better generalization than $L2$. Both loss functions are visualised for the same number of epochs.}
\label{fig:resultsCross}
\end{figure}

\textbf{Experimental setup:} The experiments have been conducted on an Intel i7 machine with a GeForce GTX 980 graphics card. The training time varies slightly between the different datasets, but in general it takes 2-3 hours to train a single ConvNet. This training time scales linearly for the case of the cascade. Furthermore, the testing time of a single ConvNet is $0.01$ seconds per image. Regarding the implementation of our algorithm, basic operations of the ConvNet such as convolution, pooling and normalization are based on MatConvNet~\cite{vedaldi15matconvnet}.

\textbf{Evaluation metrics:} We rely on the mean pixel error (MPE) to measure the performance of the ConvNets. In addition, we employ the PCP (percentage of correctly estimated parts) performance measure, which is the standard metric used in human pose estimation~\cite{Ferrari08}. We distinguish two variants of the PCP score according to the literature~\cite{pishchulin2012articulated}. In \textit{strict} PCP score, the PCP score of a limb, defined by a pair of joints, is considered correct if the distance between \textit{both} estimated joint locations and true limb joint locations is at most  $50\%$ of the length of the ground-truth limb, while the \textit{loose} PCP score considers the \textit{average} distance between the estimated joint locations and true limb joint locations. During the comparisons with other methods, we explicitly indicate which version of the PCP score is used (Table \ref{table:resultsAll}).

\subsection{Baseline Evaluation}

In the first part of the evaluation, the convergence and generalization properties of Tukey's biweight loss functions are examined using the single ConvNet $\phi(.)$ of Eq.~\eqref{ConvNetEq}, without including the cascade. We compare the results of the robust loss with $L2$ loss using the same settings and training data of PARSE~\cite{yang2013articulated}, LSP~\cite{Johnson10}, Football~\cite{kazemi2013multi} and Volleyball~\cite{belagiannis2014holistic} datasets. To that end, a 5-fold cross validation has been performed by iteratively splitting the training data of all datasets (none of the datasets includes by default a validation set), where the average results are shown in Fig.~\ref{fig:resultsCross}. Based on the results of the cross validation which is terminated by early stopping~\cite{lecun1998neural}, we have selected the number of training epochs for each dataset. After training by using all training data for each dataset, we have compared the convergence and generalization properties of Tukey's biweight and $L2$ loss functions. For that purpose, we  choose the lowest MPE of $L2$ loss and look for the epoch with the closest MPE after training with Tukey's biweight loss function. The results are summarized in Fig.~\ref{fig:resultsComparison} for each dataset. It is clear that by using Tukey's biweight loss, we obtain notably faster convergence (note that on the PARSE dataset it is $20$ times faster). This speed-up can be very useful for large-scale regression problems, where the training time usually varies from days to weeks. Besides faster convergence, we also obtain better generalization, as measured by the error in the validation set, using our robust loss (see Fig.~\ref{fig:resultsComparison}). More specifically, we achieve $12\%$ smaller MPE error using Tukey's biweight loss functions in two out of four datasets (i.e PARSE and Football), while we are  around $8\%$ better with LSP and Volleyball datasets.

\begin{figure}[t]
\centering
\setlength{\tabcolsep}{4.5pt}
\begin{tabular}{cc}
\includegraphics[scale=0.10, angle=0]{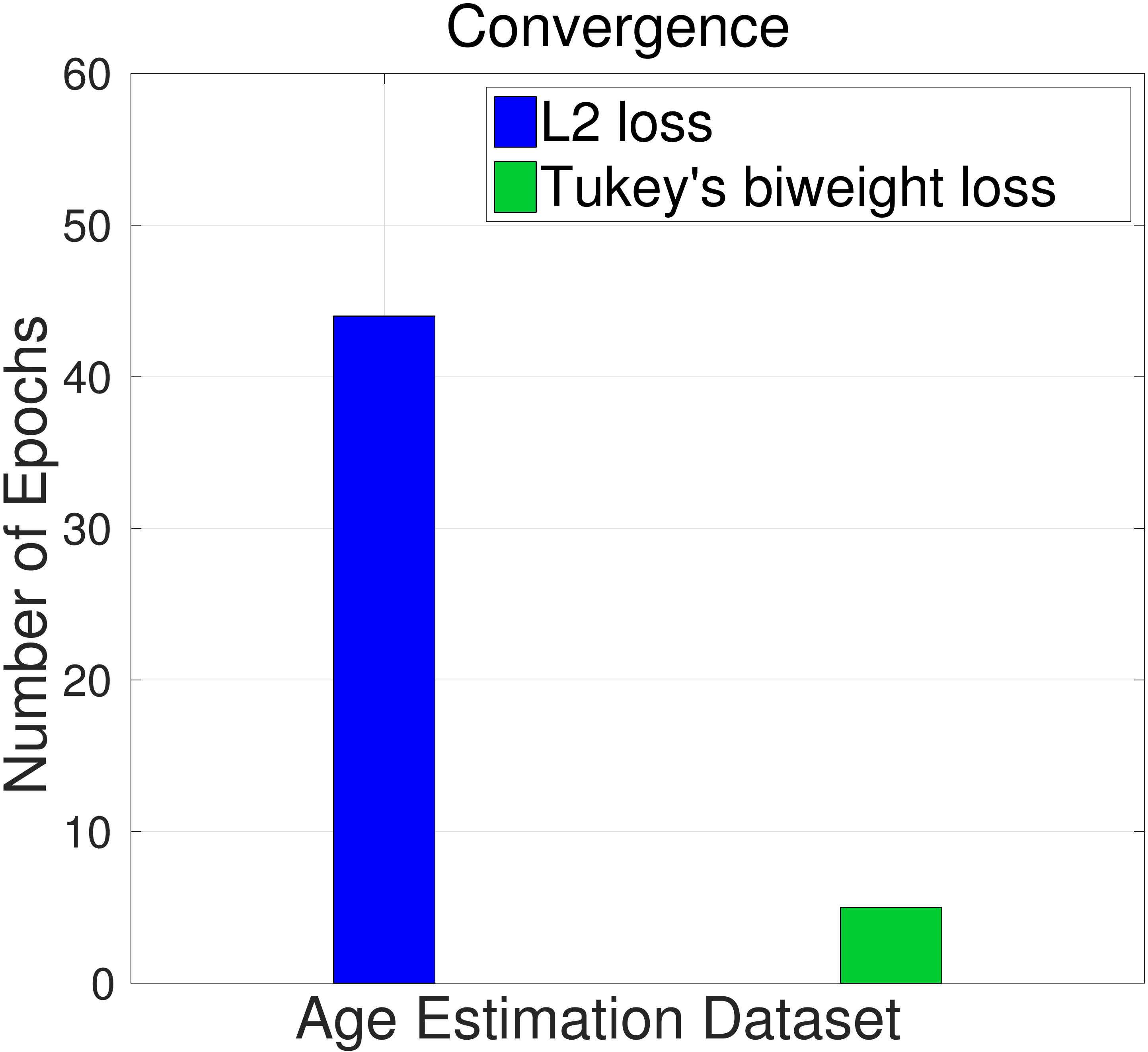}&
\includegraphics[scale=0.10, angle=0]{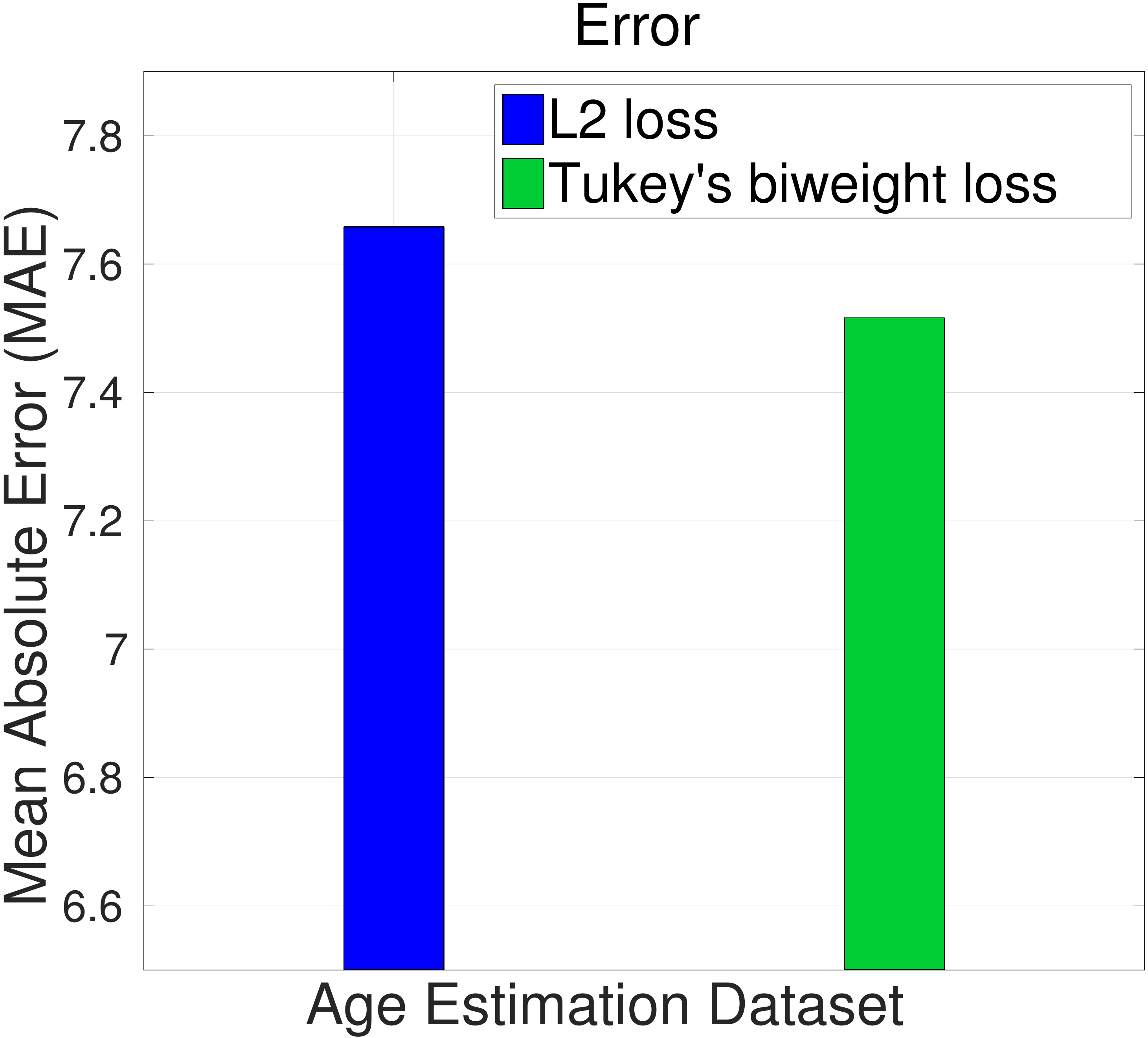}
\end{tabular}
\caption{\small {\bf Comparison of $L2$ and $Tukey's$ $biweight$ loss functions on age estimation}: Comparsion of our results (Tukey's biweight loss) with the $L2$ loss function on apparent age estimation from face images~\cite{escalera2015chalearn}. On left, the convergence of the loss functions is presented, while on the right, the mean absolute error (MAE)  in years is presented for both loss functions. For the convergence computation, we choose as reference error, the smallest error using $L2$ loss and then look for the epoch with the closest error in the training using Tukey's biweight loss.}
\label{fig:resultsComparison2}
\end{figure}

We additionally present a comparison between Tukey's biweight and \textit{L2} loss functions on age estimation from face images~\cite{escalera2015chalearn}, to demonstrate the generalization of our robust loss. In this task, we simplify the network by removing the second convolutional layer and the first fully connected layer. Moreover, we set the number of channels to 8 for all layers and the size of the remaining fully connected to 256. We randomly chose $80\%$ of the data with available annotation (2476 samples) for training and the rest for testing. In the training data, we perform augmentation and 5-fold cross validation, as in human pose estimation. Our results are summarized in Fig.~\ref{fig:resultsComparison2}, which shows faster convergence and better performance compared to \textit{L2} loss.

\subsection{Comparison with other Methods}

In this part, we evaluate our robust loss function using the coarse-to-fine model represented by the cascade of ConvNets (Fig.~\ref{fig:network}), presented in Sec.~\ref{sec:modelRef}, and compare our results with the state-of-the-art from the literature, on the four aforementioned body pose datasets (PARSE~\cite{yang2013articulated}, LSP~\cite{Johnson10}, Football~\cite{kazemi2013multi} and Volleyball~\cite{belagiannis2014holistic}). For the comparisons, we use the \textit{strict} and \textit{loose} PCP scores, depending on which evaluation metric was used by the state-of-the-art. The results are summarized in Table~\ref{table:resultsAll}, where the first row of each evaluation shows our result using a single ConvNet $\phi(.)$ of Eq.~\eqref{ConvNetEq} and the second row, the result using the cascade of ConvNets $\{ \phi^{c}(.) \}_{c=1}^C$ of Eq.~\eqref{refinedY}, where $C=3$.

\begin{figure}
\centering
\begin{tabular}{cccccc}
\includegraphics[scale=0.12, angle=-0]{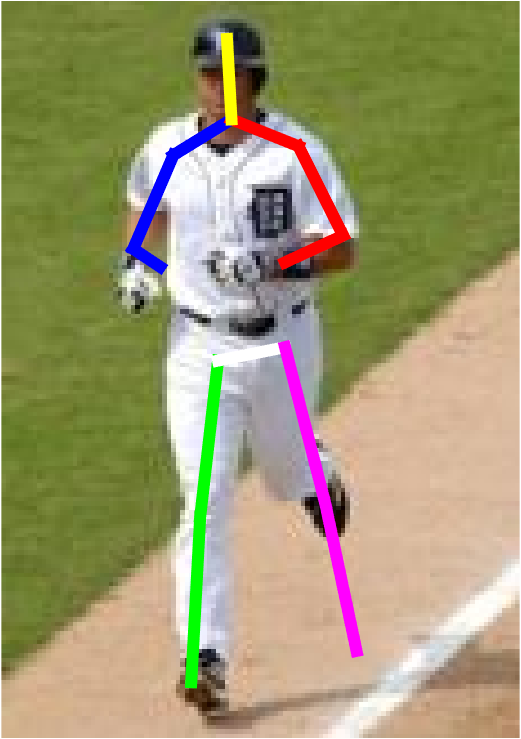}&
\includegraphics[scale=0.112, angle=-0]{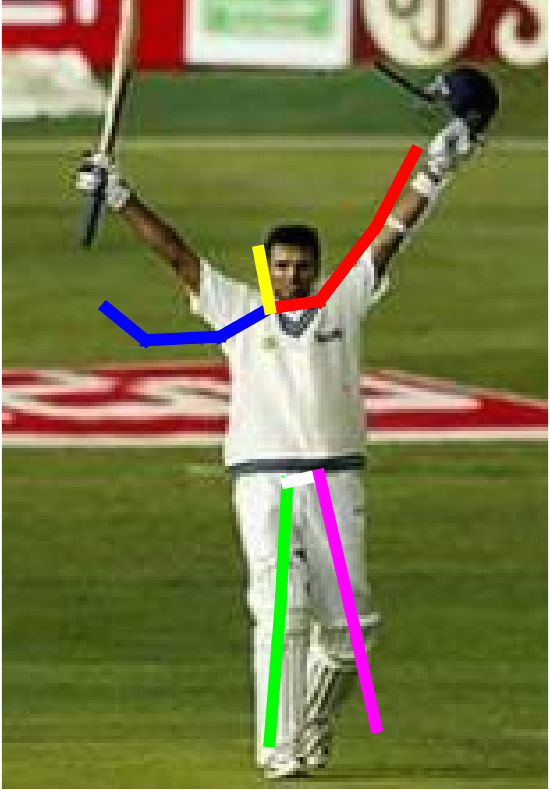}&
\includegraphics[scale=0.12, angle=-0]{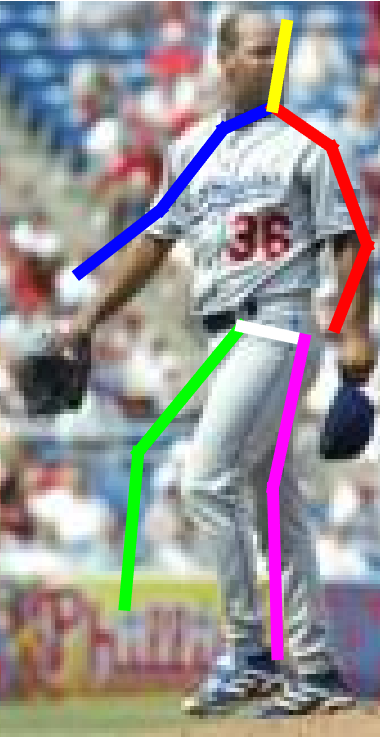}&
\includegraphics[scale=0.127, angle=-0]{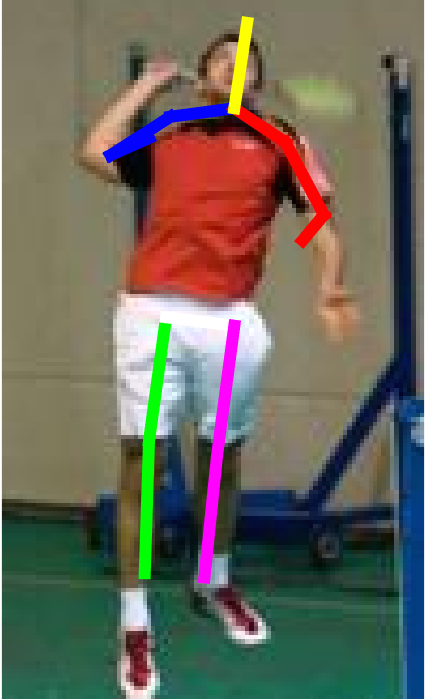}&
\includegraphics[scale=0.114, angle=-0]{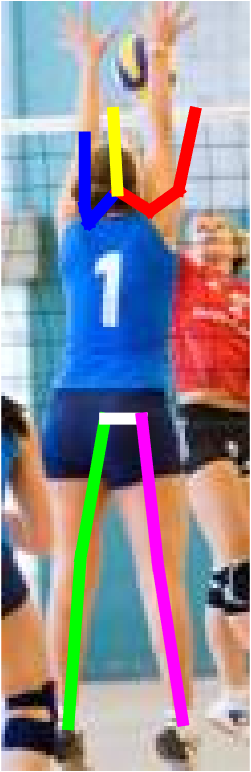}&
\includegraphics[scale=0.106, angle=-0]{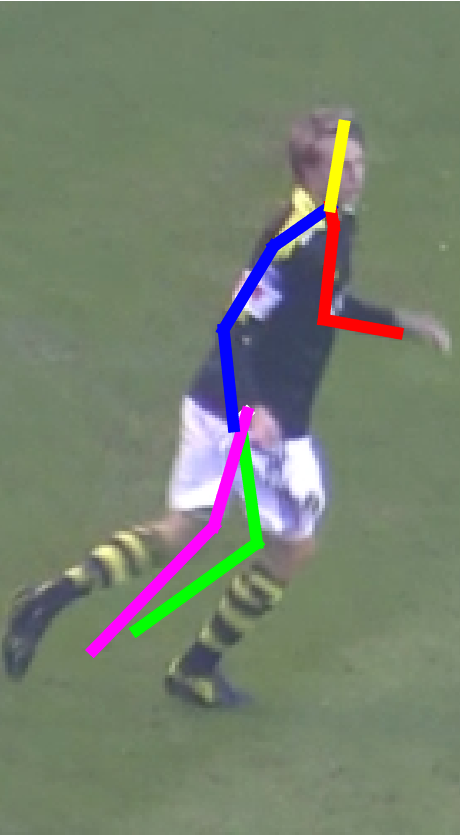}\\
\includegraphics[scale=0.12, angle=-0]{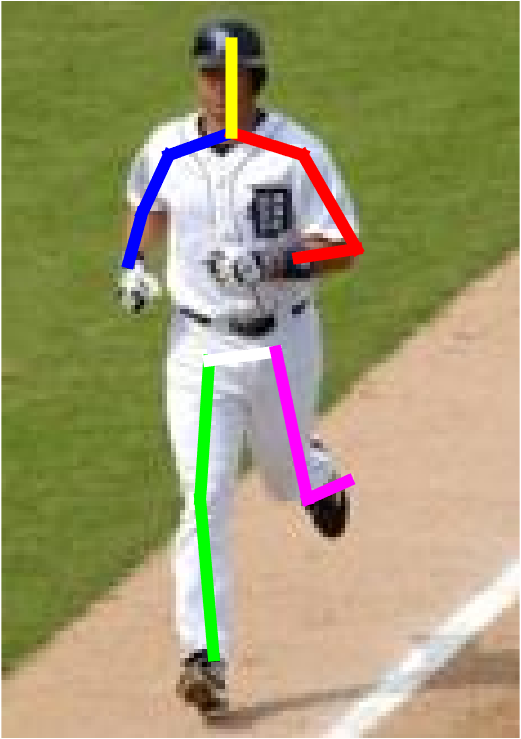}&
\includegraphics[scale=0.112, angle=-0]{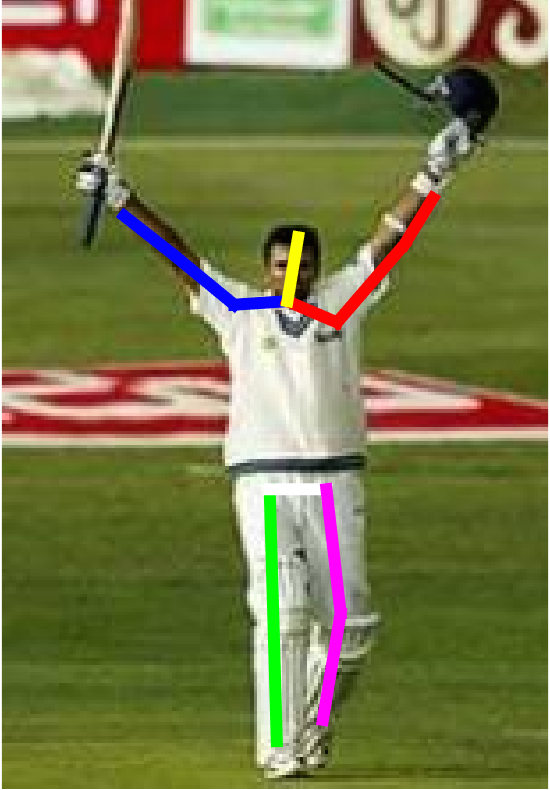}&
\includegraphics[scale=0.12, angle=-0]{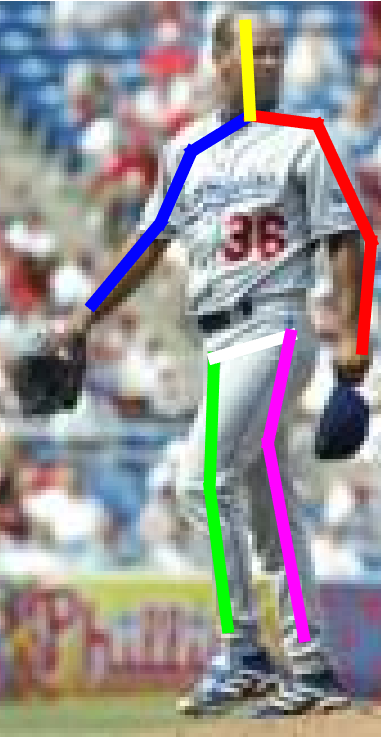}&
\includegraphics[scale=0.127, angle=-0]{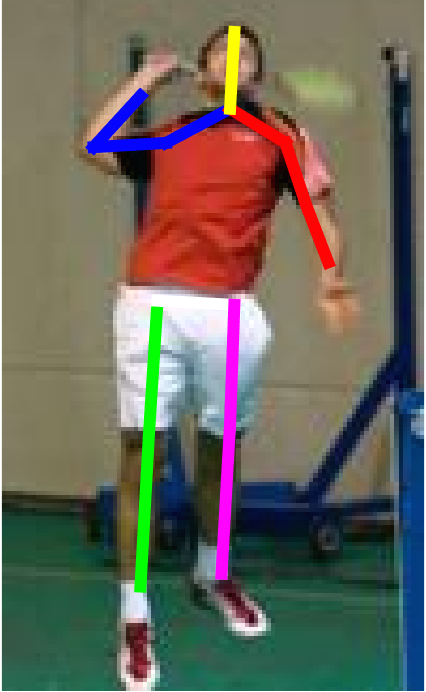}&
\includegraphics[scale=0.114, angle=-0]{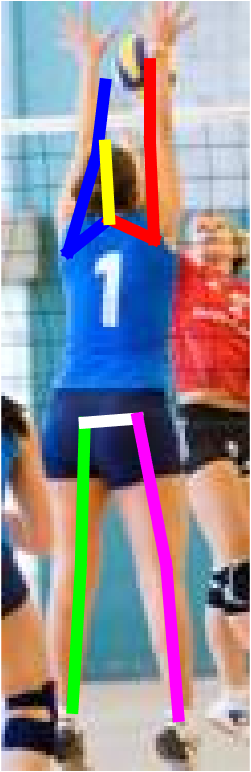}&
\includegraphics[scale=0.106, angle=-0]{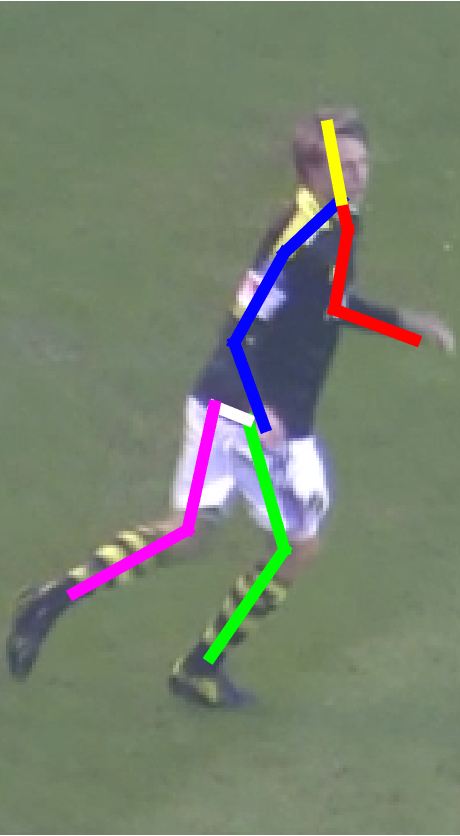}\\ \hline
 \multicolumn{3}{c|}{{\small PARSE}}  & \multicolumn{2}{c|}{{\small LSP}} &  {\footnotesize Football}\\
\end{tabular}
\caption{\small {\bf Model refinement}: Our results before (top row) and after (bottom row) the refinement with the cascade for the PARSE~\cite{yang2013articulated}, LSP~\cite{Johnson10} and Football~\cite{kazemi2013multi} datasets. We train $C=3$ ConvNets for the cascade $\{ \phi^{c}(.) \}_{c=1}^C$, based on the output of the single ConvNet $\phi(.)$.}
\label{fig:resultsRefine}
\end{figure}

\textbf{PARSE:} This is a standard dataset to assess 2D human pose estimation approaches and thus we show results from most of the current state-of-the-art, as displayed in Table~\ref{table:resultsPARSE}. While our result is $68.5\%$ for the full body regression using a single ConvNet, our final score is improved by around $5\%$ with the cascade. We achieve the best score in the full body regression as well as in most body parts. Closer to our performance is another deep learning method by Ouyang et al.~\cite{ouyang2014multi} that builds on part-based models and deep part detectors. The rest of the compared methods are also part-based, but our holistic model is simpler to implement and at the same time is shown to perform better (Fig.~\ref{fig:resultsImgParse} and \ref{fig:resultsRefine}).

\begin{figure}
\centering
\begin{tabular}{ccccc}
\includegraphics[scale=0.105, angle=-0]{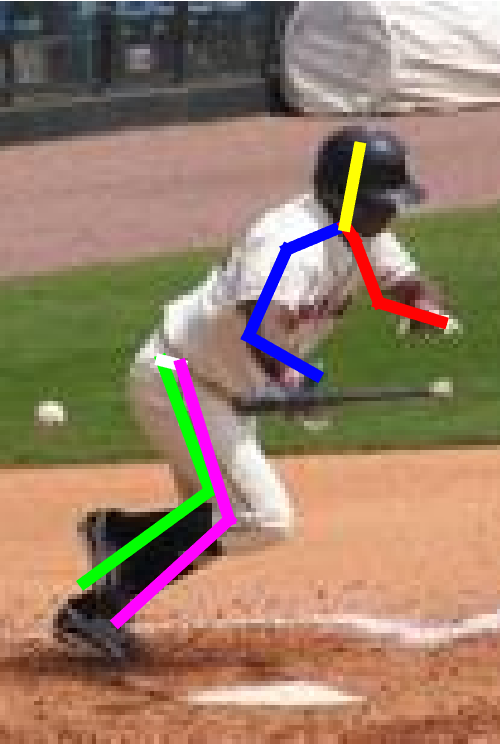}&
\includegraphics[scale=0.129, angle=-0]{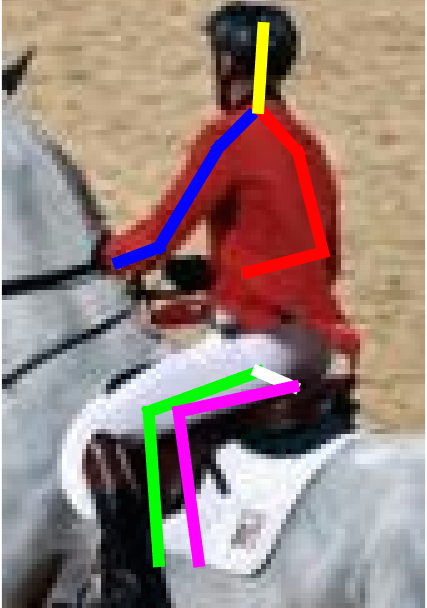}&
\includegraphics[scale=0.135, angle=-0]{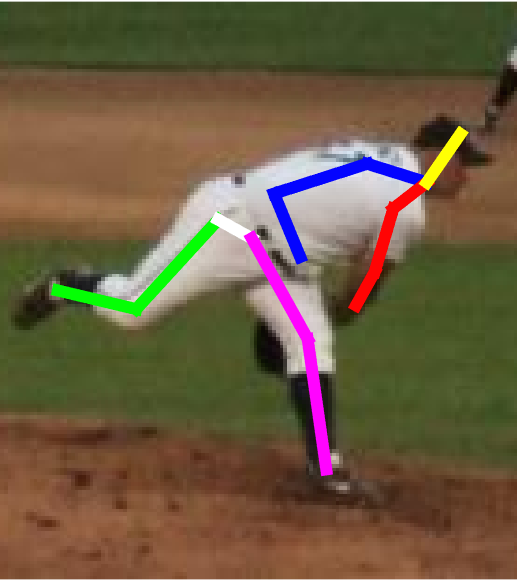}&
\includegraphics[scale=0.139, angle=-0]{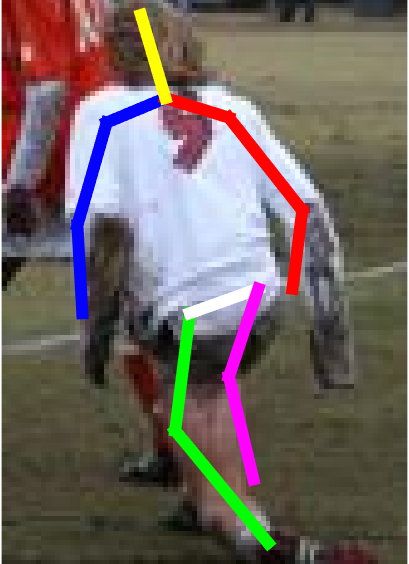}&
\includegraphics[scale=0.18, angle=90]{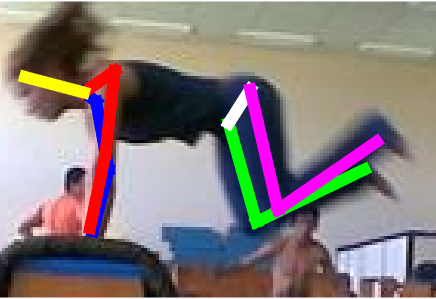}\\
\includegraphics[scale=0.115, angle=-0]{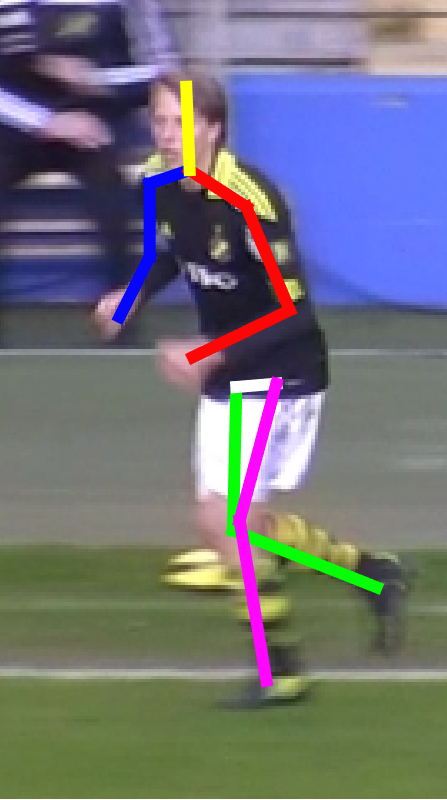}&
\includegraphics[scale=0.115, angle=-0]{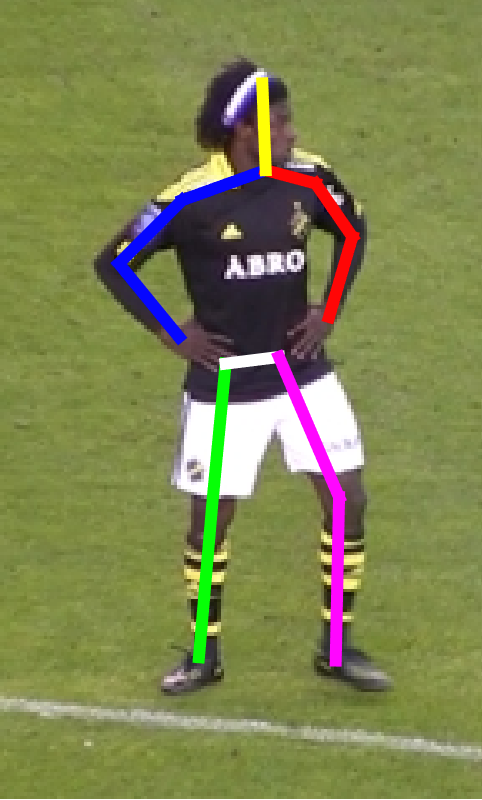}&
\includegraphics[scale=0.115, angle=-0]{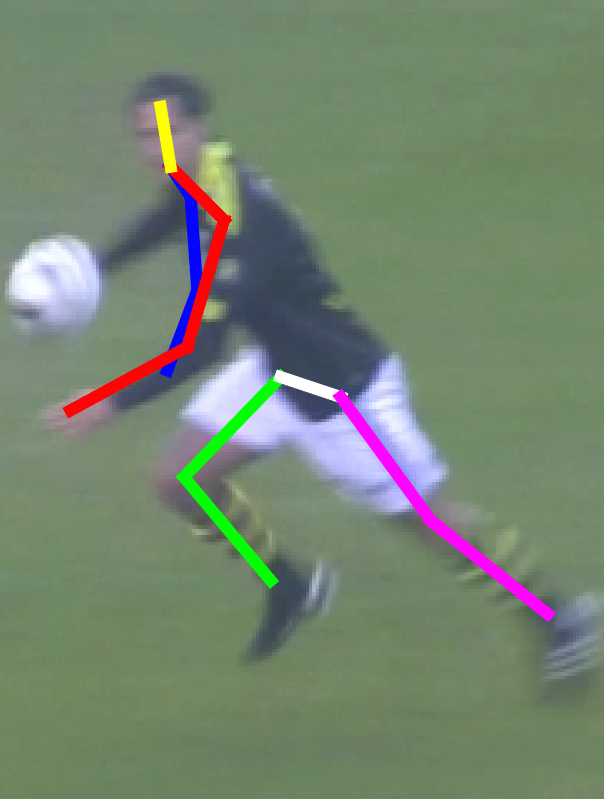}&
\includegraphics[scale=0.115, angle=-0]{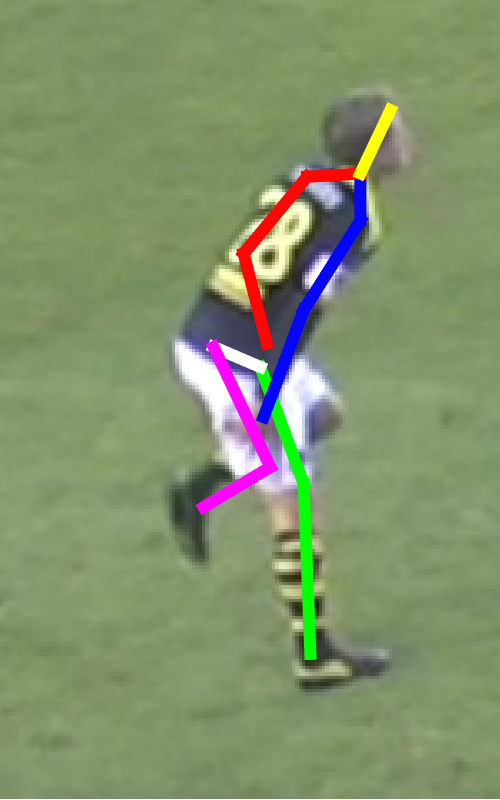}&
\includegraphics[scale=0.115, angle=-0]{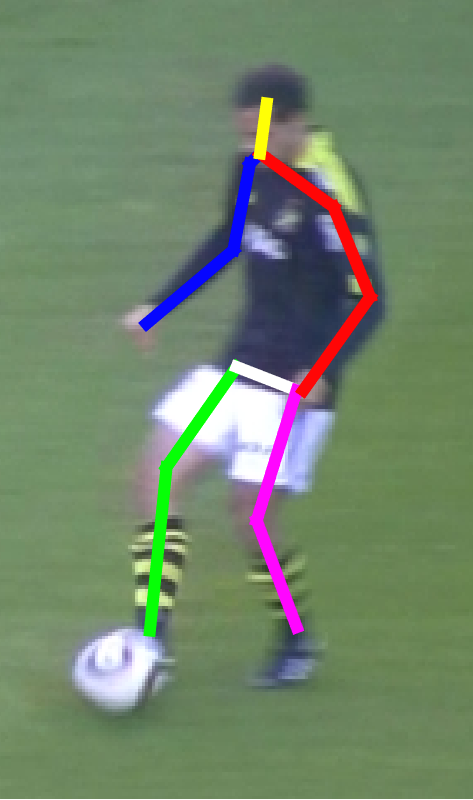}\\
\includegraphics[scale=0.115, angle=-0]{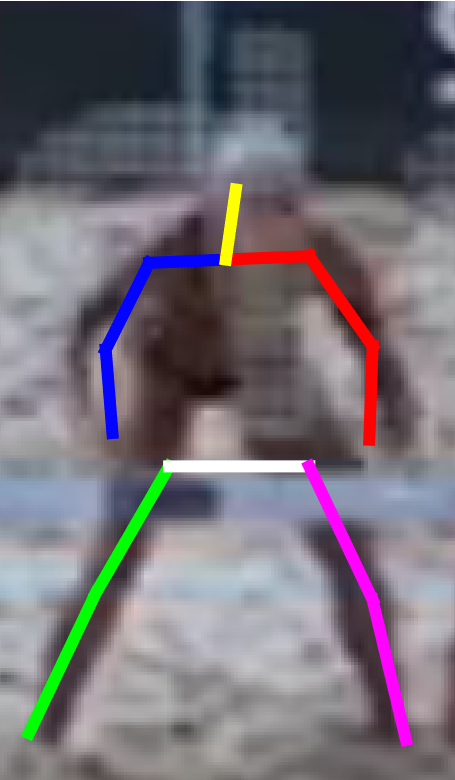}&
\includegraphics[scale=0.128, angle=-0]{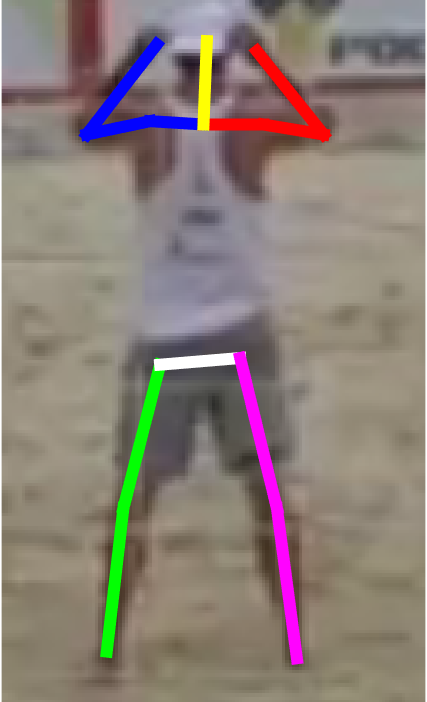}&
\includegraphics[scale=0.1185, angle=-0]{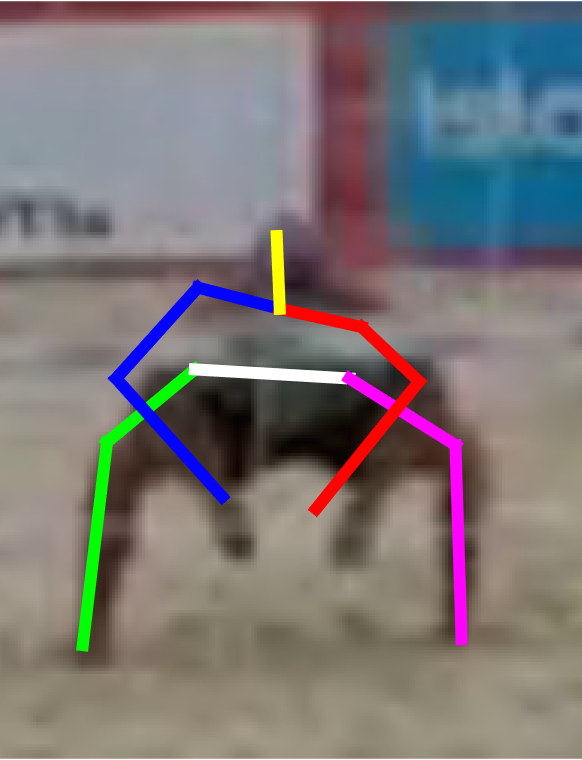}&
\includegraphics[scale=0.116, angle=-0]{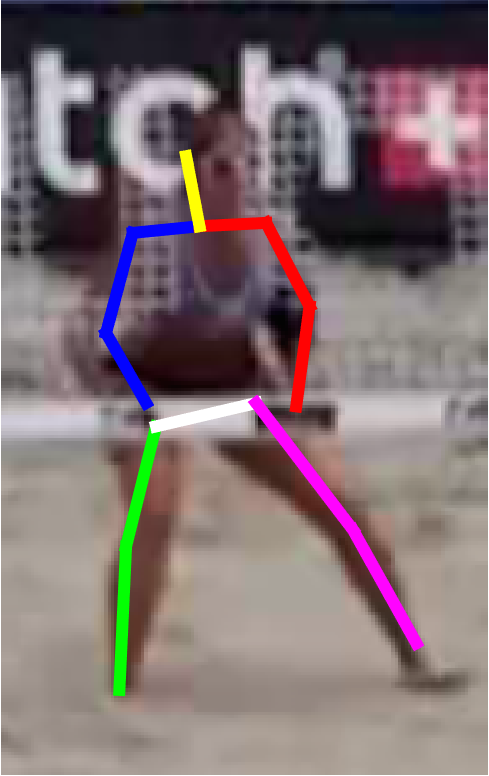}&
\includegraphics[scale=0.108, angle=-0]{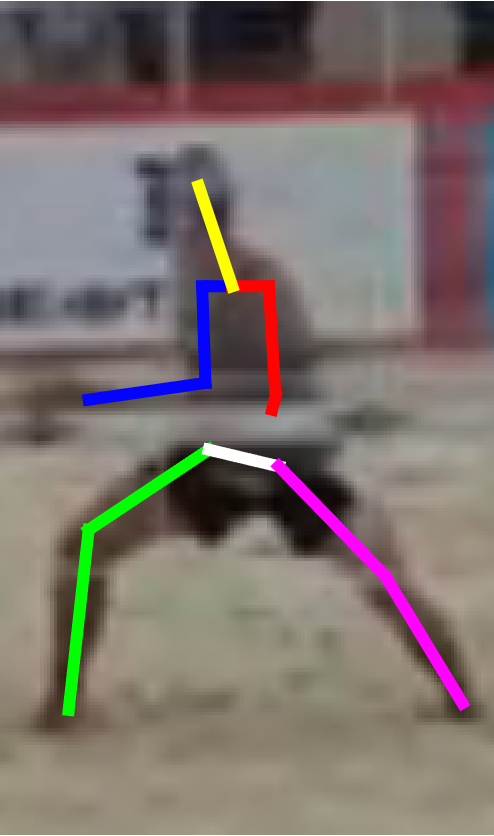}\\

\end{tabular}
\caption{\small {\bf Additional results}: Samples of our results on 2D human pose estimation are presented for the LSP~\cite{Johnson10} (first row), Football~\cite{kazemi2013multi} (second row) and Volleyball~\cite{belagiannis2014holistic} (third row) datasets.}
\label{fig:resultsAdditional}
\end{figure}

\textbf{LSP:} In LSP dataset, our approach shows a similar performance, compared to the PARSE dataset, using a single ConvNet or a cascade of ConvNets. In particular, the PCP score using one ConvNet increases again by around $5\%$ with the cascade of ConvNets, from $63.9\%$ to $68.8\%$ for the full body evaluation (Table~\ref{table:resultsLSP}). The holistic approach of Toshev et al.~\cite{toshev2014deeppose} is also a cascade of ConvNets, but it relies on $L2$ loss and different network structure. On the other hand, the Tukey's biweight loss being minimized in our network brings better results in combination with the cascade. Note also that we have used $4$ ConvNets in total for our model in comparison to the $29$ networks used by Toshev et al.~\cite{toshev2014deeppose}. Moreover, considering the performance with respect to body parts, the best PCP scores are shared between our method and the one of Chen \& Yuille~\cite{chen2014articulated}. The part-based model of Chen \& Yuille~\cite{chen2014articulated} scores best for the full body, head, torso and arms, while we obtain the best scores on the upper and lowers legs. We show some results on this dataset in Fig.~\ref{fig:resultsRefine} and \ref{fig:resultsAdditional}.
\begin{table}
\centering
\small

\begin{subtable}{1.0\linewidth}
\centering

\begin{tabular}{l|c|c|c|c|c|c|c}
               & Head & Torso & Upper & Lower & Upper & Lower & Full \\
 Method                        &  &   &              Legs &  Legs & Arms & Arm & Body \\ \hline \hline
\textit{L2} loss & 69.2 & 93.6 & 77.3 & 69.0 & 50.4 & 27.8 &61.1\\  \hline
Ours           & 78.5 & 95.6  & 82.0   & 75.6   & 61.5   & 36.6   & 68.5  \\
Ours (cascade) &\textbf{91.7}      &\textbf{98.1}       &\textbf{84.2}        &\textbf{79.3}        &66.1        &41.5        &\textbf{73.2}       \\ \hline
Andriluka et al. \cite{andriluka2009pictorial}     & 72.7 & 86.3 & 66.3 & 60.0 & 54.6 & 35.6 & 59.2 \\
Yang\&Ramanan \cite{yang2013articulated}     & 82.4 & 82.9 & 68.8 & 60.5 & 63.4 & 42.4 & 63.6 \\
Pishchulin et al. \cite{pishchulin2012articulated} & 77.6 & 90.7 & 80.0 & 70.0 & 59.3 & 37.1 & 66.1\\
Johnson et al. \cite{Johnson10}           & 76.8 & 87.6 & 74.7 & 67.1 & 67.3 & 45.8 & 67.4\\
Ouyang et al. \cite{ouyang2014multi}     & 89.3 & 89.3  & 78.0   & 72.0   & \textbf{67.8}   & \textbf{47.8}   & 71.0 
\end{tabular}
\caption{ \small {\bf PARSE Dataset} The evaluation metric on PARSE dataset \cite{yang2013articulated} is the \textit{strict} PCP score.}
\label{table:resultsPARSE}
\end{subtable}

\begin{subtable}{1.0\linewidth}
\centering

\begin{tabular}{l|c|c|c|c|c|c|c}
               & Head & Torso & Upper & Lower & Upper & Lower & Full \\
 Method                        &  &   &              Legs &  Legs & Arms & Arm & Body \\ \hline \hline
\textit{L2} loss & 68.2 & 90.4 & 77.0 & 67.7 & 51.9 & 26.6 & 60.5\\  \hline
Ours           & 72.0 & 91.5  & 78.0       & 71.2     & 56.8       & 31.9       & 63.9      \\
Ours (cascade) & 83.2 & 92.0  &\textbf{79.9}       &\textbf{74.3}     & 61.3       & 40.3       & 68.8      \\ \hline
Toshev et al.  \cite{toshev2014deeppose}      & -    & -     & 77.0       & 71.0     & 56.0       & 38.0       & -         \\
Kiefel\&Gehler \cite{kiefel2014human}    & 78.3 & 84.3  & 74.5       & 67.6     & 54.1       & 28.3       & 61.2      \\
Yang\&Ramanan \cite{yang2013articulated}       & 79.3 & 82.9  & 70.3       & 67.0     & 56.0       & 39.8       & 62.8      \\
Pishchulin et al.  \cite{pishchulin2012articulated}   & 85.1 & 88.7  & 78.9       & 73.2     & 61.8       & 45.0       & 69.2      \\
Ouyang et al.  \cite{ouyang2014multi}      & 83.1 & 85.8  & 76.5       & 72.2     & 63.3       & 46.6       & 68.6      \\
Chen\&Yuille   \cite{chen2014articulated}      &\textbf{87.8} &\textbf{92.7}  & 77.0       & 69.2     & \textbf{69.2}       & \textbf{55.4}       &\textbf{75.0}     
\end{tabular}
\caption{ \small {\bf LSP Dataset} The evaluation metric on LSP dataset \cite{Johnson10} is the \textit{strict} PCP score.}
\label{table:resultsLSP}
\end{subtable}

\begin{subtable}{1.0\linewidth}
\centering

\begin{tabular}{l|c|c|c|c|c|c|c}
               & Head & Torso & Upper & Lower & Upper & Lower & Full \\
 Method                        &  &   &              Legs &  Legs & Arms & Arm & Body \\ \hline \hline
\textit{L2} loss & 96.7 & 99.4 & 98.8 & 97.8& 95.4 & 84.5  & 94.8\\  \hline
Ours           & 97.1 & 99.7  & 99.0   & 98.1   & 96.2   & 87.1   & 95.8  \\
Ours (cascade) &\textbf{98.3}      &\textbf{99.7}       &\textbf{99.0}        &\textbf{98.1}        &\textbf{96.6}        &\textbf{88.7}        &\textbf{96.3}       \\ \hline
Yang\&Ramanan \cite{yang2013articulated}        & 97.0      & 99.0       & 94.0        & 80.0       &92.0        & 66.0       &86.0       \\
Kazemi et al.  \cite{kazemi2013multi}  & 96.0      & 98.0      &97.0        & 88.0       &  93.0      &   71.0     &  89.0 
\end{tabular}
\caption{ \small {\bf Football Dataset} The evaluation metric on Football dataset \cite{kazemi2013multi} is the \textit{loose} PCPscore.}
\label{table:resultsFoot}
\end{subtable}

\begin{subtable}{1.0\linewidth}
\centering

\begin{tabular}{l|c|c|c|c|c|c|c}
               & Head & Torso & Upper & Lower & Upper & Lower & Full \\
 Method                        &  &   &              Legs &  Legs & Arms & Arm & Body \\ \hline \hline
\textit{L2} loss & 89.3 & 96.6 & 90.4 & 91.8 & 68.2 & 50.1  & 78.7\\  \hline
Ours           & 90.4 & \textbf{97.1}  &\textbf{86.4}   &\textbf{95.8}   &74.0   & 58.3   & \textbf{81.7}  \\
Ours (cascade) &89.0      &95.8       &84.2        &94.0        &\textbf{74.2}        &\textbf{58.9}        &81.0       \\ \hline
Yang\&Ramanan \cite{yang2013articulated}   & 76.1      & 80.5      &52.4        & 70.5       &  40.7      &   33.7     &  56.0     \\
Belagiannis et al. \cite{belagiannis2014holistic}  & \textbf{97.5}      & 81.4       & 65.1        & 81.2       &54.4        & 19.3       &60.2       \\
\end{tabular}
\caption{ \small {\bf Volleyball Dataset} The evaluation metric on Volleyball dataset \cite{belagiannis2014holistic} is the \textit{loose} PCP score.}
\label{table:resultsVolley}
\end{subtable}
\caption{ \small {\bf  Comparison with related work}: We compare our results (Tukey's biweight loss) using one ConvNet (second row) and the cascade of ConvNets (third row). We also provide the scores of the training using the $L2$ loss (first row). The scores of the other methods are the ones reported in their original papers.}
\label{table:resultsAll}
\end{table}

\textbf{Football:} This dataset has been introduced by Kazemi et al.~\cite{kazemi2013multi} for estimating the 2D pose of football players. Our results (Table~\ref{table:resultsFoot}) using one ConvNet are almost optimal (with a PCP score of $95.8\%$)  and thus the improvement using the cascade is smaller. However, it is worth noting that effective refinements are achieved with the use of the cascade of ConvNets, as demonstrated in Fig.~\ref{fig:resultsRefine} and \ref{fig:resultsAdditional}.

\textbf{Volleyball:} Similar to the Football dataset~\cite{kazemi2013multi}, our results on the Volleyball dataset are already quite competitive using one ConvNet (Table~\ref{table:resultsVolley}), with a PCP score of  $81.7\%$. On this dataset, the refinement step has a negative impact to our results (Table~\ref{table:resultsVolley}). We attribute this behaviour to the interpolation results of the cropped images, since the original images have low resolution (last row of Fig.~\ref{fig:resultsAdditional}).

\section{Conclusion}
We have introduced \textit{Tukey's biweight} loss function for the robust optimization of ConvNets in regression-based problems. Using 2D human pose estimation and age estimation from face images as testbed, we have empirically shown that optimizing with this loss function, which is robust to outliers, results in faster convergence and better generalization compared to the standard $L2$ loss, which is a common loss function used in regression problems. We have also introduced a cascade of ConvNets that improves the accuracy of the localization in 2D human pose estimation. The combination of our robust loss function with the cascade of ConvNets produces comparable or better results than the state-of-the-art methods in four public human pose estimation datasets.

\section*{Acknowledgments}

This work was partly funded by DFG (``Advanced Learning for Tracking
\& Detection in Medical Workflow Analysis'') and TUM - Institute for Advanced Study (German Excellence Initiative - FP7 Grant 291763). G. Carneiro thanks the Alexander von Humboldt Foundation (Fellowship for Experienced Researchers) \& the Australian Research Council Centre of Excellence for Robotic Vision (project number CE140100016) for providing funds for this work.

{\small
\bibliographystyle{ieee}
\bibliography{mybib}
}

\end{document}